\documentclass[11pt]{article}
\usepackage[utf8]{inputenc}
\usepackage[T1]{fontenc}
\usepackage[a4paper,margin=1in]{geometry}
\usepackage{booktabs}
\usepackage{amsmath}
\usepackage{amssymb}
\usepackage{hyperref}
\usepackage{microtype}
\usepackage[section]{placeins}
\usepackage{etoolbox}
\AtBeginEnvironment{table}{\footnotesize}

\title{ECG Biometrics with ArcFace-Inception: External Validation on MIMIC and HEEDB}
\author{Scagnetto A.}
\date{}

\begin{document}
\maketitle

\begin{abstract}
\textbf{Introduction:} ECG biometrics has been evaluated predominantly on small cohorts and on comparisons between temporally close recordings, conditions that structurally simplify the identification problem and produce performance metrics that are difficult to transfer to more demanding operational settings. It therefore remains an open question how identification systems behave in large-scale scenarios, with galleries containing thousands of identities and temporal gaps between examinations on the order of years. This work addresses both dimensions of the problem by adopting a unified, reproducible evaluation protocol with external multi-domain validation.

\textbf{Methods:} The model is based on an Inception-v1 1D backbone trained with ArcFace loss on 12-lead ECGs, producing 512-dimensional L2-normalized embeddings compared through cosine similarity. Training was conducted on the internal ASUGI corpus (164{,}440 ECGs, 53{,}079 patients), built from 1{,}140{,}500 raw recordings through criteria of quality, patient traceability, and acquisition-device homogeneity. External evaluation was performed on datasets derived from MIMIC-IV-ECG (231{,}329 ECGs, 63{,}895 patients) and HEEDB (385{,}079 ECGs, 118{,}756 patients) according to four protocols: general comparability (closed-set leave-one-out), scale analysis (gallery 500--7{,}000 identities), temporal stress test (1--5 year gap at constant gallery size), and post-hoc reranking (score normalization, diffusion, query expansion).

\textbf{Results:} Under the general comparability protocol, the system reaches Rank@1 values of 0.9506 on ASUGI-DB, 0.8291 on MIMIC-GC, and 0.6884 on HEEDB-GC, with TAR@FAR$=10^{-3}$ of 0.9675, 0.8188, and 0.7210, respectively. The scale analysis shows a monotonic decay of Rank@1 as gallery size increases and a systematic recovery as the number of examinations per patient grows. In the temporal stress test, Rank@1 decreases from 0.7853 (1 year) to 0.6433 (5 years) on MIMIC and from 0.6864 to 0.5560 on HEEDB. In the focused AS-norm reranking analysis on HEEDB-RR, the best observed configuration reaches Rank@1=0.8005 compared with the 0.7765 baseline. Confidence-score analysis shows substantial discriminative separation in all domains, with $\Delta = \bar{c}_{y=1} - \bar{c}_{y=0}$ equal to 0.4882 on ASUGI-DB, 0.3731 on MIMIC-GC, and 0.4204 on HEEDB-GC; at $\tau = 0.90$, selective prediction covers 88.05\%, 58.89\%, and 35.73\% of queries, respectively, with Err@0.90 of 0.14\%, 1.66\%, and 3.31\%.

\textbf{Discussion:} Although numerically lower than many values reported in the literature, the performance reported here refers to a substantially harder identification problem: galleries of tens of thousands of identities, multi-center domain shift, and temporal distances up to five years. In this setting, the proposed model maintains robust discriminative separation, with a predictable and quantifiable performance degradation as scale and temporal distance increase. The results should be interpreted as evidence of robustness in a large-scale closed-set stress test with external validation, rather than as a complete demonstration of readiness for clinical deployment.
\end{abstract}

\noindent\textbf{Keywords:} ECG biometrics; large-scale identification; ArcFace; temporal stress test; confidence-score calibration; domain shift; MIMIC-IV-ECG; HEEDB

\section{Introduction}

The electrocardiogram (ECG) is a physiological signal that encodes morphological and temporal information characteristic of each individual, while still being influenced by clinical status, acquisition conditions, and inter-session variability. Despite this variability, numerous experimental studies and systematic literature reviews document the existence of an identity component sufficiently stable to support biometric verification (1:1) and identification (1:N) tasks. Historical works and more recent surveys consistently report a statistically and operationally significant separation between intra-subject (genuine) and inter-subject (impostor) comparisons, with identity information persisting even across months or years, albeit with degraded performance relative to single-session scenarios \cite{odinaka2012,donida2019,fratini2015,fatemian2009,dasilva2014}.

The introduction of deep learning models and embedding-based representations has accelerated the development of ECG biometric systems, expanding the set of available architectures and improving performance on standard benchmarks \cite{merone2017,pinto2018,melzi2023}. However, the literature presents several structural weaknesses that limit comparability and operational interpretation. The first is methodological: heterogeneity in experimental settings -- evaluation protocols, split criteria, and reported metrics -- makes direct comparison across studies difficult and obscures system behavior at operationally realistic security thresholds \cite{odinaka2012,donida2019,fratini2015}. The second is scale: most published works rely on cohorts of only a few hundred or a few thousand identities, leaving open the question of how systems behave in operational regimes with clinically relevant gallery sizes \cite{odinaka2012,donida2019,pinto2018}.

This work addresses both weaknesses. Methodologically, it adopts a unified, reproducible evaluation protocol -- leave-one-out with Rank@K and TAR@FAR metrics -- conceived as a large-scale closed-set stress test with external multi-domain validation. In terms of scale, the internal ASUGI-DB dataset comprises more than 164{,}000 ECGs from over 53{,}000 patients, with an internal test set of about 52{,}000 ECGs from more than 20{,}000 identities, already an order of magnitude larger than the cohorts most commonly used in the literature. As a further distinguishing element, the study includes external validation on public corpora that are even larger: MIMIC-IV-ECG \cite{mimic4wdb_010} contributes 231{,}329 ECGs from 63{,}895 patients, while HEEDB \cite{koscova2025heedb,heedb_50,goldberger2000} contributes 385{,}079 ECGs from 118{,}756 patients, making it the largest public ECG dataset with known identities used so far in a biometric setting. The combination of large-scale training on internal data and external validation on even larger cohorts, with controlled domain shift and temporal distances up to five years, defines a particularly severe experimental setting within the ECG biometrics literature, while remaining distinct from an open-set deployment scenario.

\section{Materials and methods}

\subsection{Materials: datasets}

This study uses three ECG corpora: one internal (\textbf{ASUGI}) and two public (\textbf{MIMIC-IV-ECG} and \textbf{HEEDB}). Derived datasets are denoted by corpus name plus protocol suffix: \textbf{ASUGI-DB} (training, validation, and internal test), \textbf{MIMIC-GC} and \textbf{HEEDB-GC} (general comparability), \textbf{MIMIC-TST} and \textbf{HEEDB-TST} (temporal stress test), \textbf{HEEDB-scale} (scale analysis), and \textbf{HEEDB-RR} (reranking). Corpus names denote raw sources; suffixed names denote derived experimental datasets. A compact overview is provided in the Supplementary Materials file \texttt{materiali\_aggiuntivi-EN.tex} (Table~S1).

\paragraph{ASUGI domain.}
The internal corpus is owned by ASUGI -- Azienda Sanitaria Universitaria Giuliano-Isontina (Trieste, Italy). In raw form it contains 1{,}140{,}500 ECG recordings acquired between 2006-01-01 and 2020-12-31 during routine clinical activity. Based on the \texttt{PatientID} field, 687{,}147 ECGs from 191{,}354 patients had valid identifiers and were therefore eligible for biometric analysis; 55{,}867 ECGs had invalid non-missing identifiers and 397{,}486 had missing identifiers and were excluded. Only the valid-\texttt{PatientID} subset was considered for dataset construction (Section~\ref{sec:pipeline}).

Recordings were acquired with the \textbf{ELI250} device (Mortara Instrument, Milwaukee, WI, USA), a digital 12-lead electrocardiograph used in routine hospital diagnostics. The device acquires the 12 standard leads simultaneously at 1000 Hz and exports integer-valued clinical XML signals with sufficient dynamic range for downstream analysis. It also provides automated fiducial measurements (intervals, axes, amplitudes), which were used for outlier filtering in the internal domain. Additional technical specifications are reported in the Supplementary Materials (Table~S3).

After applying the filtering protocol described in the next section, the cohort used in the study contains 164{,}440 ECGs from 53{,}079 patients.

\paragraph{MIMIC-IV-ECG domain.}
MIMIC-IV-ECG: Diagnostic Electrocardiogram Matched Subset \cite{mimic4wdb_010} is an open-access database developed by the MIT Laboratory for Computational Physiology in collaboration with Beth Israel Deaconess Medical Center. Version 1.0 contains about 800{,}000 diagnostic 12-lead ECGs from nearly 160{,}000 patients, acquired during 2008--2019 at 500 Hz for 10-second recordings. All recordings come from a single center but span emergency, inpatient, and outpatient settings; the subset used here (231{,}329 ECGs, 63{,}895 patients) was built without restricting the clinical setting. Data are distributed in WFDB format through PhysioNet.

\paragraph{HEEDB domain.}
The Harvard-Emory ECG Database (HEEDB) \cite{koscova2025heedb,heedb_50,goldberger2000} is a large open-access 12-lead ECG database. Version 5.0 includes about 11.67 million recordings from about 2.17 million patients, collected in routine clinical care from the early 1980s onward. Recordings are 10 seconds long and sampled at 250 Hz or 500 Hz depending on the site and acquisition period. The dataset includes demographic metadata and automated diagnostic annotations.

This study uses only the \textbf{i0006} subset (Emory University Hospital, Atlanta). After filtering, this subset contains 385{,}079 ECGs from 118{,}756 patients.

\subsection{Dataset construction}
\label{sec:pipeline}
All derived datasets were constructed from the three corpora through the filtering pipeline summarized below.

\paragraph{Common pipeline across all datasets.}
A valid \texttt{PatientID} is required to construct genuine and impostor comparisons. The common filtering criteria applied to all corpora are summarized in Table~\ref{tab:pipeline_comune}.

\begin{table}[ht]
\centering
\caption{Common filtering pipeline applied to all datasets.}
\label{tab:pipeline_comune}
\small
\begin{tabular}{cp{5.5cm}p{6.5cm}}
\toprule
Step & Criterion & Rationale \\
\midrule
1 & Valid \texttt{PatientID} & Fundamental requirement for constructing genuine and impostor pairs \\[4pt]
2 & At least 2 ECGs per patient; minimum distance of 30 days between consecutive examinations & Ensures intra-subject comparisons with real temporal variability \\[4pt]
3 & Maximum cap of 10 ECGs per patient & Limits imbalance due to patients with very frequent examinations \\
\bottomrule
\end{tabular}
\end{table}

\paragraph{ASUGI-DB.}
In addition to the common pipeline, ASUGI-DB applied two dataset-specific filters: (i) outlier removal based on fiducial ECG features (full ranges in the Supplementary Materials, Table~S10) and (ii) device harmonization to ELI250 recordings only. These steps were not replicated on MIMIC and HEEDB because comparable fiducial and device metadata are not available in the public releases. Cross-domain comparisons must therefore be interpreted under asymmetric data curation. The attrition flow is reported in Table~\ref{tab:asugidb_attrition}. ASUGI-DB was used for training, validation, and internal test.

\begin{table}[ht]
\centering
\caption{ASUGI-DB construction flow: sample size at each filtering step.}
\label{tab:asugidb_attrition}
\small
\begin{tabular}{clrr}
\toprule
Step & Description & ECGs & Patients \\
\midrule
1 & Initial base with valid \texttt{PatientID} & 687{,}147 & 191{,}355 \\
2 & Multi-exam filter + temporal constraint ($>$30 days between consecutive examinations) & 395{,}095 & 92{,}523 \\
3 & Outlier cleaning on fiducial features + maximum 10 ECGs/patient & 168{,}999 & 54{,}218 \\
4 & Device harmonization (ELI250 only) & 164{,}440 & 53{,}079 \\
\bottomrule
\end{tabular}
\par\medskip\footnotesize\textit{Note: attrition flow of the internal ASUGI-DB dataset; steps 1--2 correspond to the common pipeline, whereas steps 3--4 are specific to the internal dataset.}
\end{table}

Three patient-disjoint ASUGI configurations (db1000/db1001/db1002) were derived from ASUGI-DB. They differ only in the minimum number of ECGs per patient required in the training split (3/4/5, respectively), whereas validation and test require at least 2 ECGs per patient (Supplementary Materials, Table~S2). Consistent with previous results \cite{scagnetto2026ecg}, db1001 was adopted as the reference configuration (Table~\ref{tab:db1001_splits}).

\begin{table}[ht]
\centering
\caption{Patient-disjoint split of training dataset db1001 (ASUGI-DB).}
\label{tab:db1001_splits}
\small
\begin{tabular}{lrrrrr}
\toprule
Split & ECGs & Patients & Min ECG/pat. & Mean ECG/pat. & Max ECG/pat. \\
\midrule
Train      & 60{,}153 & 11{,}647 & 4 & 5.16 & 10 \\
Validation & 51{,}988 & 20{,}716 & 2 & 2.51 & 10 \\
Test       & 52{,}299 & 20{,}716 & 2 & 2.52 & 10 \\
\bottomrule
\end{tabular}
\par\medskip\footnotesize\textit{Note: db1001 requires at least 4 ECGs per patient in training and at least 2 in validation and test. It is the reference configuration selected based on the results of the previous study \cite{scagnetto2026ecg}.}
\end{table}

\paragraph{Protocol-specific datasets.}
The MIMIC and HEEDB corpora were used only for external validation under four protocols: \textit{general comparability} (MIMIC-GC and HEEDB-GC), \textit{scale analysis} (HEEDB-scale), \textit{temporal stress test} (MIMIC-TST and HEEDB-TST), and \textit{reranking} (HEEDB-RR).

\textbf{MIMIC-GC} and \textbf{HEEDB-GC} were constructed by applying the common pipeline to the respective corpora; for HEEDB-GC, only 500 Hz recordings were retained to match the training signal. The resulting pools comprise 231{,}329 ECGs from 63{,}895 patients for MIMIC-GC and 385{,}079 ECGs from 118{,}756 patients for HEEDB-GC.

\textbf{HEEDB-scale} is a stratified sample of HEEDB-GC in which gallery size (500--7{,}000 patients) and number of examinations per patient (2--7) are varied on a controlled grid. These limits represent the largest jointly feasible range, because larger galleries and stricter exams-per-patient requirements progressively reduce the admissible patient pool.

\textbf{HEEDB-RR} is built from HEEDB-GC through proportional stratified sampling over the distribution of ECGs per patient (target $P=10{,}000$, seed 42), retaining all ECGs of selected patients. This size was chosen to preserve the source distribution while keeping reranking sweeps computationally tractable; the agreement between HEEDB-RR and HEEDB-GC is reported in Table~\ref{tab:heedb_rr_moments}. On HEEDB-RR, baseline embedding and all reranking methods are evaluated at fixed gallery size.

\begin{table}[ht]
\centering
\caption{Comparison of the moments of the \#ECG/patient distribution between HEEDB-GC (full) and HEEDB-RR ($P=10{,}000$, seed 42).}
\label{tab:heedb_rr_moments}
\small
\begin{tabular}{lrrrrr}
\toprule
Dataset & Mean ($\mu_1$) & Variance ($\mu_2$) & Skewness ($\gamma_1$) & Kurtosis ($\beta_2$) & Excess kurtosis \\
\midrule
HEEDB-GC (full)          & 3.2426 & 2.9533 & 1.7000 & 5.6226 & 2.6226 \\
HEEDB-RR ($P=10{,}000$)  & 3.2428 & 2.9552 & 1.7006 & 5.6246 & 2.6246 \\
\midrule
$|\Delta|$               & 0.0002 & 0.0019 & 0.0005 & 0.0020 & 0.0020 \\
\bottomrule
\end{tabular}
\par\medskip\footnotesize\textit{Note: absolute differences between the full corpus and the sample are negligible, confirming that proportional stratified sampling preserves the shape of the source distribution.}
\end{table}

\textbf{MIMIC-TST} and \textbf{HEEDB-TST} are ECG-pair datasets built from MIMIC-GC and HEEDB-GC with target gaps of 1, 2, 3, 4, and 5 years (tolerance $\pm$3 months) and at most one pair per patient for each target. To remove confounding due to year-specific gallery size, sample size was fixed to the smallest subset (5-year target): $P=14{,}521$ for MIMIC-TST and $P=14{,}763$ for HEEDB-TST.

\subsection{Study protocols}

\subsubsection{General comparability}
The primary evaluation uses leave-one-out closed-set identification: each ECG is used as a query and compared with all remaining ECGs using cosine similarity between L2-normalized embeddings. Rank@K and TAR@FAR are reported for FAR$\in\{10^{-3},10^{-4}\}$.

\subsubsection{Temporal Stress Test}
The temporal stress test quantifies performance degradation as the gap between paired ECGs increases. For each dataset (MIMIC-TST and HEEDB-TST), the query ECG is compared with its paired genuine ECG and with ECGs from all other patients in the gallery. Evaluation is performed separately for each temporal target (1--5 years) at constant gallery size ($P=14{,}521$ for MIMIC-TST and $P=14{,}763$ for HEEDB-TST). Rank@K and TAR@FAR are reported.

\subsubsection{Scale analysis}
On HEEDB-scale, controlled combinations of \texttt{gallery\_size} and \texttt{exams\_per\_patient} quantify degradation with gallery growth and recovery as more examinations per identity become available.

\subsubsection{Reranking}

Reranking is a post-hoc stage that updates the top-$K$ shortlist produced by embedding matching using contextual gallery information not available in purely pairwise comparison. All methods operate on L2-normalized embeddings with cosine similarity as the base score and do not require a second feature-extraction pass.

\paragraph{Algorithms used.}

\subparagraph{Best-of-$K$ (max-score fusion).}
For each candidate identity in the shortlist, the final score is the maximum over the available comparisons with samples from the same identity. This score-level ``max'' fusion is used as the second-stage baseline \cite{kittler1998}.

\subparagraph{Score normalization: Z-norm, T-norm, S-norm, AS-norm, C-norm.}
Classical biometric score-normalization methods \cite{auckenthaler2000,poh2012} calibrate scores with respect to an impostor cohort: Z-norm operates on the gallery side, T-norm on the query side, S-norm combines the two symmetrically, and AS-norm uses an adaptive cohort of top impostors. A combined internal variant (C-norm) was also tested. In all cases, ranking changes indirectly through local score calibration.

\subparagraph{Diffusion reranking (graph-based).}
Diffusion reranking propagates the initial score along the local connectivity of a gallery k-NN graph, thereby incorporating higher-order similarities according to manifold-diffusion methods used in retrieval \cite{donoser2013,iscen2017}.

\subparagraph{AQE/$\alpha$QE (query expansion).}
Average Query Expansion (AQE) updates the query by aggregating the top-$K$ neighbors and reruns retrieval. In the $\alpha$QE variant, neighbor contributions are weighted by similarity raised to the power $\alpha$, reducing the impact of weak neighbors \cite{chum2007,radenovic2018}.

\paragraph{Configuration sets}

\subparagraph{Experiment 1 -- internal cohort (HEEDB-RR).}
On HEEDB-RR, we evaluated the baseline leave-one-out setup, diffusion reranking over a grid of $K$, local\_$k$, $\alpha$, and iteration values, score normalization with cohort internal to the gallery (S-norm, Z-norm, C-norm), AS-norm with embedding cache, and $\alpha$QE over a grid of $K$ and $\alpha$.

\subparagraph{Experiment 2 -- external HEEDB cohort.}
Using the same HEEDB-RR evaluation gallery, we tested AS-norm, S-norm, Z-norm, C-norm, and T-norm with a cohort extracted from an external HEEDB pool non-overlapping with the gallery. Cohort size varied in $\{100, 250, 500, 1000, 2000, 3000\}$ patients (seed 42).

\subparagraph{Experiment 3 -- external ASUGI-DB cohort.}
The setup is identical to Experiment 2, except that the normalization cohort is extracted from ASUGI-DB instead of HEEDB, allowing cross-domain portability of normalization to be assessed.

\paragraph{Confidence calibrator and evaluation metrics.}
The confidence framework is applied to both GC and HEEDB-RR. In both settings, the main calibrator is a two-feature logistic regression with input $x=[s_1,s_2]^\top$, feature-wise standardization, and sigmoidal output:
\[
\tilde{x} = \frac{x-\mu}{\sigma}, \qquad
z = w_1\tilde{x}_1 + w_2\tilde{x}_2 + b, \qquad
p_{\text{top1\_correct}} = \sigma(z) = \frac{1}{1 + e^{-z}},
\]
where $\mu$ and $\sigma$ are estimated on the calibration subset only, $\sigma(\cdot)$ is the sigmoid function, and $(w_1,w_2,b)$ are obtained by optimizing cross-entropy with L2 regularization. The target variable is $y_i \in \{0,1\}$, with $y_i = 1$ if the rank-1 candidate matches the query identity and $y_i = 0$ otherwise.

The calibrator structure is identical in the two settings; only the definition of $s_1$ and $s_2$ changes. In GC, they correspond to the first two ECG-level scores in the leave-one-out ranking; in HEEDB-RR they are the first two scores after AS-norm reranking. In both cases, calibration and evaluation subsets are generated through patient-disjoint stratified splits with fixed seed.

The main text reports only Brier score, Expected Calibration Error (ECE), Coverage@$\tau$, Error@$\tau$, and the separation statistics $\bar{c}_{y=1}$, $\bar{c}_{y=0}$, and $\Delta$. Additional metrics, the full reranking-confidence tables, GC confidence intervals, and alternative calibrators are reported in the Supplementary Materials (Tables~S20--S31).

\subsection{Neural network}

The model consists of an \textbf{Inception-v1 1D} backbone \cite{szegedy2015,szegedy2016} that produces 512-dimensional L2-normalized embeddings and an \textbf{ArcFace} classification head \cite{deng2019} used only during training. At inference time, the head is removed and cosine similarity is computed directly between embeddings.

\paragraph{Inception-v1 1D architecture used.}
The backbone \cite{szegedy2015,szegedy2016} comprises an input convolutional stem, three Inception blocks separated by downsampling layers, and a projection head toward the embedding space. Each Inception block processes the signal through four parallel branches with different receptive fields -- a pointwise $1{\times}1$ convolution, two paths $1{\times}1 \to k{\times}1$ with $k \in \{3,5\}$, and a max-pooling branch -- whose outputs are concatenated. The complete stage-wise structure is reported in Table~\ref{tab:inception_arch}.

\begin{table}[ht]
\centering
\caption{Architecture of the Inception-v1 1D backbone (input: 12 channels $\times$ T samples).}
\label{tab:inception_arch}
\small
\begin{tabular}{llp{6.5cm}r}
\toprule
Stage & Type & Configuration & Out ch. \\
\midrule
Stem         & Conv1D-BN-ReLU $\times$2   & kernel 7, stride 1; 12$\to$64$\to$64                                        & 64  \\
Inception 1  & 4 branches $+$ concat $+$ dropout & $1{\times}1$ /\ $1{\times}1{\to}3{\times}1$ /\ $1{\times}1{\to}5{\times}1$ /\ MaxPool${\to}1{\times}1$; $p=0.05$ & 128 \\
Downsample 1 & Conv1D stride 2            & kernel 3                                                                     & 128 \\
Inception 2  & 4 branches $+$ concat $+$ dropout & same structure; $p=0.05$                                                    & 192 \\
Downsample 2 & Conv1D stride 2            & kernel 3                                                                     & 192 \\
Inception 3  & 4 branches $+$ concat $+$ dropout & same structure; $p=0.05$                                                    & 256 \\
Embedding    & AvgPool $+$ Linear $+$ L2  & global average pooling; 256$\to$512; L2 normalization                        & 512 \\
\bottomrule
\end{tabular}
\par\medskip\footnotesize\textit{Note: the ArcFace head (used only in training) is attached after the linear layer, before L2 normalization.}
\end{table}

\paragraph{Why ArcFace was chosen.}
ArcFace was selected based on previous experiments on the same ECG corpus \cite{scagnetto2026ecg}, where it outperformed pairwise and triplet metric-learning objectives in both Rank@1 and TAR@FAR and showed more stable optimization in large-gallery settings. This is consistent with its formulation: embeddings and class weights are normalized on the hypersphere, and an additive angular margin is optimized in the same cosine geometry used at inference, reducing the mismatch between training objective and retrieval metric.

\paragraph{ArcFace head and training.}
\paragraph{ArcFace head and training.}
The ArcFace head applies an additive angular margin to normalized logits. The number of classes equals the number of identities in the training set (11{,}647 patients for db1001). In the main protocol, the parameters are scale $s=32.0$, margin warmup from 0.3 to 0.5 during the first 15 epochs, AdamW optimization, and batch size 64. Training was not run for a fixed number of epochs; instead, early stopping was applied based on validation Rank@1, with patience of 20 epochs and a minimum improvement threshold of 0.001. The total number of training epochs therefore varied across runs as a function of validation performance. The implementation is based on PyTorch with CUDA and was trained on a single NVIDIA RTX 4090 GPU (24\,GB VRAM).

\paragraph{ArcFace loss (formulation).}
Following the original formulation \cite{deng2019}, ArcFace optimizes cross-entropy on normalized embeddings and class weights with an additive angular margin on the correct class. In this study, the head is used only during training; at test time, comparison is performed directly between L2-normalized embeddings.

\paragraph{Signal preprocessing and normalization.}
Each ECG recording is a $12 \times 10{,}000$ matrix (12 leads, 10 seconds at 1000 Hz), which is costly to process during training. To evaluate the trade-off between temporal resolution and computational cost, training was repeated at 1000 Hz, 500 Hz, and 250 Hz. Downsampling was preceded by an anti-aliasing windowed FIR filter (Hamming, 101 taps, cutoff $=0.45 \times f_\text{target}$), followed by integer-step decimation.

The 250 Hz configuration produced a marked performance loss, whereas the difference between 1000 Hz and 500 Hz was minimal. For this reason, 500 Hz was selected as the reference configuration because it reduces computational cost while remaining fully compatible with MIMIC and HEEDB, both natively sampled at 500 Hz.

In all evaluation settings, signals were normalized per channel (z-score) using mean and standard deviation estimated on the training split of the selected run, thus avoiding leakage from validation and external test data.

\section{Results}

Unless otherwise stated, Rank@K metrics were computed in a closed-set leave-one-out regime.

\subsection{Overall performance.}

\begin{table}[ht]
\centering
\caption{Overall performance (general comparability, closed-set) on ASUGI-DB, MIMIC-GC, and HEEDB-GC.}
\label{tab:global_three_datasets}
\begin{tabular}{lccccc}
\toprule
Dataset & Rank@1 & Rank@5 & Rank@10 & TAR@FAR=1e-3 & TAR@FAR=1e-4 \\
\midrule
ASUGI-DB  & 0.9506 & 0.9670 & 0.9718 & 0.9675 & 0.9448 \\
MIMIC-GC  & 0.8291 & 0.8768 & 0.8916 & 0.8188 & 0.7079 \\
HEEDB-GC  & 0.6884 & 0.7593 & 0.7807 & 0.7210 & 0.6002 \\
\bottomrule
\end{tabular}
\par\medskip\footnotesize\textit{Note: compact comparison across the three general-comparability datasets under the same evaluation protocol. TAR@FAR columns are included for comparability with prior work; a full verification-threshold analysis is outside the scope of this study.}
\end{table}

Performance decreases from ASUGI-DB to MIMIC-GC to HEEDB-GC, consistent with increasing heterogeneity in the external domains. The magnitude of this gradient should also be interpreted in light of asymmetric data curation, because fiducial filtering and device harmonization were applied only to the internal dataset.

\subsection{Confidence score analysis -- General comparability.}

Table~\ref{tab:confidence_gc} summarizes confidence metrics under the general comparability protocol, using top-1 correctness as the binary reference variable. The reported confidence is the calibrated probability $p_{\text{top1\_correct}}$ produced by the main calibrator \texttt{conf\_lgr2} from the first and second ECG-level scores in the retrieval ranking.

\begin{table}[ht]
\centering
\caption{Confidence metrics under the general comparability protocol.
$\bar c_{y=1}$ and $\bar c_{y=0}$ denote the mean calibrated confidence for correct and incorrect top-1 identifications;
$\Delta = \bar c_{y=1} - \bar c_{y=0}$.
Brier and ECE measure probabilistic calibration.
Cov@0.90 and Err@0.90 report coverage and conditional error at operating point $\tau=0.90$.}
\label{tab:confidence_gc}
\footnotesize
\setlength{\tabcolsep}{4pt}
\begin{tabular}{lccccccc}
\toprule
& \multicolumn{3}{c}{Discriminability} & \multicolumn{2}{c}{Calibration} & \multicolumn{2}{c}{Selective prediction} \\
\cmidrule(lr){2-4}\cmidrule(lr){5-6}\cmidrule(lr){7-8}
Dataset & $\bar c_{y=1}$ & $\bar c_{y=0}$ & $\Delta$ & Brier & ECE & Cov@0.90 & Err@0.90 \\
\midrule
ASUGI-DB & 0.9702 & 0.4820 & 0.4882 & 0.0207 & 0.0185 & 0.8805 & 0.0014 \\
MIMIC-GC & 0.8916 & 0.5185 & 0.3731 & 0.0849 & 0.0340 & 0.5889 & 0.0166 \\
HEEDB-GC & 0.8173 & 0.3969 & 0.4204 & 0.1185 & 0.0286 & 0.3573 & 0.0331 \\
\bottomrule
\end{tabular}
\par\medskip\footnotesize\textit{Note: Acc@0.5 is not reported in the table; the values are 0.9713, 0.8756, and 0.8366 for ASUGI-DB, MIMIC-GC, and HEEDB-GC, respectively.}
\end{table}

Confidence remains discriminative in all three domains, with $\Delta = \bar{c}_{y=1} - \bar{c}_{y=0}$ equal to 0.4882 on ASUGI-DB, 0.3731 on MIMIC-GC, and 0.4204 on HEEDB-GC. Thus, correct identifications receive substantially higher confidence than errors even when retrieval performance declines out of domain.

Operationally, high-confidence coverage decreases from ASUGI-DB to MIMIC-GC to HEEDB-GC (Cov@0.90: 0.8805, 0.5889, 0.3573), while residual error increases (Err@0.90: 0.0014, 0.0166, 0.0331). Discrimination is therefore preserved, but high-confidence decisions become less frequent and less reliable in the external domains. The 95\% confidence intervals ($n=5$ seeds) for GC aggregate metrics are reported in the Supplementary Materials (Tables~S30--S31).

\subsection{HEEDB-scale.}

\begin{table}[ht]
\centering
\caption{HEEDB-scale -- Rank@1 as a function of gallery size (500--7000) and number of examinations per patient (2--7). Mean over seeds 0--5.}
\label{tab:heedb_rank1_grid}
\footnotesize
\begin{tabular}{rrrrrrrr}
\toprule
Gallery & Ex=2 & Ex=3 & Ex=4 & Ex=5 & Ex=6 & Ex=7 & Mean \\
\midrule
500  & 0.7517 & 0.8568 & 0.8967 & 0.9083 & 0.9205 & 0.9224 & 0.8761 \\
1000 & 0.7286 & 0.8354 & 0.8798 & 0.8940 & 0.9057 & 0.9116 & 0.8592 \\
1500 & 0.7173 & 0.8255 & 0.8699 & 0.8856 & 0.8973 & 0.9052 & 0.8501 \\
2000 & 0.7063 & 0.8162 & 0.8652 & 0.8809 & 0.8934 & 0.9011 & 0.8438 \\
2500 & 0.6959 & 0.8108 & 0.8596 & 0.8759 & 0.8891 & 0.8978 & 0.8382 \\
3000 & 0.6864 & 0.8052 & 0.8549 & 0.8716 & 0.8854 & 0.8940 & 0.8329 \\
3500 & 0.6796 & 0.8001 & 0.8506 & 0.8699 & 0.8822 & 0.8909 & 0.8289 \\
4000 & 0.6737 & 0.7962 & 0.8467 & 0.8668 & 0.8792 & 0.8878 & 0.8251 \\
4500 & 0.6689 & 0.7937 & 0.8440 & 0.8644 & 0.8766 & 0.8858 & 0.8222 \\
5000 & 0.6652 & 0.7901 & 0.8408 & 0.8624 & 0.8745 & 0.8839 & 0.8195 \\
5500 & 0.6627 & 0.7868 & 0.8378 & 0.8600 & 0.8724 & 0.8823 & 0.8170 \\
6000 & 0.6601 & 0.7846 & 0.8356 & 0.8580 & 0.8702 & 0.8805 & 0.8148 \\
6500 & 0.6580 & 0.7820 & 0.8336 & 0.8564 & 0.8681 & 0.8785 & 0.8128 \\
7000 & 0.6550 & 0.7798 & 0.8308 & 0.8549 & 0.8668 & 0.8768 & 0.8107 \\
\midrule
Mean & 0.6864 & 0.8045 & 0.8533 & 0.8721 & 0.8844 & 0.8928 & 0.8322 \\
\bottomrule
\end{tabular}
\par\medskip\footnotesize\textit{Note: Rank@1 decreases with gallery size and increases with the number of examinations per patient. Rank@5, Rank@10, Rank@100, and Rank@200 are reported in the Supplementary Materials (Tables~S4--S7); 95\% confidence intervals ($n=6$ seeds) are reported in Table~S27.}
\end{table}

Table~\ref{tab:heedb_rank1_grid} shows a monotonic decrease in Rank@1 as gallery size increases and a monotonic increase as more examinations per patient are available. The mean across examination counts decreases from 0.8761 at gallery size 500 to 0.8107 at 7{,}000, whereas the mean across gallery sizes increases from 0.6864 with two examinations per patient to 0.8928 with seven. Complete Rank@5, Rank@10, Rank@100, and Rank@200 tables are reported in the Supplementary Materials (Tables~S4--S7); the corresponding 95\% confidence intervals are reported in Table~S27.

\subsection{MIMIC-TST and HEEDB-TST}

\begin{table}[ht]
\centering
\caption{Temporal Stress Test at constant gallery size: Rank@1 on MIMIC-TST and HEEDB-TST ($\pm$3 months).}
\label{tab:tst_const_rank1}
\small
\begin{tabular}{crr}
\toprule
Gap & \shortstack{MIMIC-TST \\ ($P=14{,}521$)} & \shortstack{HEEDB-TST \\ ($P=14{,}763$)} \\
\midrule
1y & 0.7853 & 0.6864 \\
2y & 0.7473 & 0.6452 \\
3y & 0.7157 & 0.6047 \\
4y & 0.6834 & 0.5776 \\
5y & 0.6433 & 0.5560 \\
\bottomrule
\end{tabular}
\par\medskip\footnotesize\textit{Note: complete metrics (Rank@5, Rank@10, TAR@FAR) are reported in the Supplementary Materials (Tables~S11--S12).}
\end{table}

At constant gallery size, Rank@1 decreases monotonically with temporal gap in both external datasets: from 0.7853 to 0.6433 on MIMIC and from 0.6864 to 0.5560 on HEEDB between 1 and 5 years. Complete Rank@5, Rank@10, and TAR@FAR metrics are reported in the Supplementary Materials (Tables~S11--S12), whereas the corresponding variable-gallery version is reported in Tables~S8--S9.

\subsection{HEEDB-RR: reranking}

Table~\ref{tab:rr_summary} reports the configurations retained for the compact cross-method comparison on HEEDB-RR. The corresponding best-per-algorithm table with explicit configuration identifiers is reported in Table~S18, complete reranking grids and the configuration-code legend in Tables~S13--S17, and the confidence tables in Tables~S20--S26.

Score-normalization methods (AS-norm, S-norm, Z-norm, C-norm) were evaluated both with gallery-internal and external cohorts. T-norm was evaluated only with an external cohort, whereas diffusion and AQE/$\alpha$QE were evaluated only in the no-external-cohort setting because they do not use impostor cohorts.

\begin{table}[ht]
\centering
\caption{Compact per-algorithm summary for HEEDB-RR: essential retrieval and confidence-calibration metrics for the configurations retained in the main cross-method comparison. \textbf{Bold}: best value per column (baseline excluded).}
\label{tab:rr_summary}
\footnotesize
\setlength{\tabcolsep}{4pt}
\begin{tabular}{lrrrrrrrr}
\toprule
Algorithm & Rank@1 & Rank@5 & Rank@10 & Acc@0.5 & Brier & ECE & Cov@0.90 & Err@0.90 \\
\midrule
baseline        & 0.7765 & 0.8304 & 0.8471 & --     & --     & --     & --     & -- \\
\textbf{asnorm} & \textbf{0.7931} & \textbf{0.8417} & \textbf{0.8573} & \textbf{0.9216} & \textbf{0.0572} & 0.0319 & \textbf{0.6429} & \textbf{0.0031} \\
snorm           & 0.7817 & 0.8358 & 0.8537 & 0.8633 & 0.1038 & 0.0349 & 0.4614 & 0.0362 \\
znorm           & 0.7765 & 0.8304 & 0.8471 & 0.8508 & 0.1122 & 0.0492 & 0.4073 & 0.0411 \\
tnorm           & 0.7644 & 0.8238 & 0.8416 & 0.8538 & 0.1039 & 0.0377 & 0.4817 & 0.0327 \\
$\alpha$QE      & 0.7626 & 0.8174 & 0.8259 & 0.8609 & 0.0975 & 0.0439 & 0.5128 & 0.0201 \\
cnorm           & 0.7583 & 0.8293 & 0.8501 & 0.8794 & 0.0866 & \textbf{0.0290} & 0.5309 & 0.0148 \\
diffusion       & 0.2567 & 0.4528 & 0.5443 & 0.8568 & 0.1107 & 0.0655 & 0.0000 & NaN \\
\bottomrule
\end{tabular}
\par\medskip\footnotesize
\textit{Configuration identifiers:}
asnorm~\texttt{internal\_K400\_N200\_scan2000};
snorm and cnorm~\texttt{ext\_heedb\_size3000\_K400\_N100\_seed42};
znorm~\texttt{ext\_heedb\_size100\_K400\_N100\_seed42};
tnorm~\texttt{ext\_internal\_size3000\_K400\_N100\_seed42};
$\alpha$QE~\texttt{K3\_a3.0};
diffusion~\texttt{K200\_lk15\_a0.950\_it8}.
Brier and ECE measure calibration; Cov@0.90 and Err@0.90 measure selective prediction. Full hyperparameter results are reported in the Supplementary Materials (Tables~S13--S19 and~S20--S26).
\end{table}

In the compact comparison, AS-norm provides the best retrieval performance among reranking methods (Rank@1 = 0.7931, Rank@5 = 0.8417, Rank@10 = 0.8573), improving over the baseline Rank@1 of 0.7765. C-norm yields the lowest ECE (0.0290), whereas diffusion strongly degrades retrieval performance. These results identify AS-norm as the most effective reranking family on HEEDB-RR.

We therefore performed a focused AS-norm analysis by varying only parameter $N$ (\texttt{asnorm\_top\_n}). Sensitivity checks indicated that, at fixed $N$, the remaining hyperparameters had limited impact on Rank@K and mainly affected calibration metrics.

\begin{table}[ht]
\centering
\caption{AS-norm focus as a function of $N$ (HEEDB-RR, no external cohort).}
\label{tab:asnorm_focus_n}
\setlength{\tabcolsep}{3.5pt}
\begin{tabular}{rrrrrrrrr}
\toprule
$N$ & Rank@1 & Rank@5 & Rank@10 & Acc@0.5 & Brier & ECE & Cov@0.90 & Err@0.90 \\
\midrule
1   & 0.7943 & 0.8387 & 0.8552 & 0.9623 & \textbf{0.0391} & 0.0690 & \textbf{0.6573} & \textbf{0.0000} \\
2   & 0.7977 & 0.8326 & 0.8472 & 0.7974 & 0.1580 & \textbf{0.0332} & 0.0182 & \textbf{0.0000} \\
3   & 0.8003 & 0.8371 & 0.8527 & 0.9499 & 0.0653 & 0.1403 & 0.5124 & \textbf{0.0000} \\
4   & \textbf{0.8005} & 0.8386 & 0.8541 & \textbf{0.9682} & 0.0536 & 0.1051 & 0.5710 & \textbf{0.0000} \\
5   & 0.7999 & 0.8389 & 0.8543 & 0.9653 & 0.0482 & 0.0970 & 0.6009 & \textbf{0.0000} \\
6   & 0.7995 & 0.8393 & 0.8549 & 0.9648 & 0.0460 & 0.0885 & 0.6140 & \textbf{0.0000} \\
7   & 0.7996 & 0.8398 & 0.8552 & 0.9078 & 0.0840 & 0.0592 & 0.4151 & \textbf{0.0000} \\
8   & 0.7987 & 0.8402 & 0.8556 & 0.9071 & 0.0837 & 0.0570 & 0.4177 & \textbf{0.0000} \\
9   & 0.7988 & 0.8405 & 0.8560 & 0.9633 & 0.0443 & 0.0772 & 0.6293 & \textbf{0.0000} \\
10  & 0.7987 & \textbf{0.8407} & \textbf{0.8561} & 0.9626 & 0.0443 & 0.0744 & 0.6319 & \textbf{0.0000} \\
\bottomrule
\end{tabular}
\par\medskip\footnotesize\textit{Setup: HEEDB-RR, run \texttt{20260208\_123000\_db1001} (best epoch 99). The full AS-norm focus table, including \texttt{conf\_mean}, \texttt{conf\_mean\_y1}, and \texttt{conf\_mean\_y0}, is reported in the Supplementary Materials (Table~S19). The corresponding 95\% confidence intervals across seeds are reported in Table~S28. Within this focused no-external-cohort sweep, Rank@1 peaks at 0.8005 for $N=4$, whereas Rank@5 and Rank@10 peak at $N=10$.}
\end{table}

\FloatBarrier

\section{Discussion}

The results indicate that ECG identity information remains measurable at large scale, but with strong dependence on domain and protocol. The discussion can therefore be organized along four axes: cross-dataset comparability, temporal robustness, reranking, and confidence quality.

\paragraph{General comparability (ASUGI-DB, MIMIC-GC, HEEDB-GC).}
Table~\ref{tab:global_three_datasets} shows a consistent performance gradient, with ASUGI-DB as the most favorable domain, MIMIC-GC intermediate, and HEEDB-GC the most severe. This pattern is consistent with increasing heterogeneity outside the internal domain, but it also reflects asymmetric curation because only ASUGI-DB benefits from fiducial filtering and device harmonization. The result should therefore be read as evidence of robustness under increasingly difficult closed-set conditions, not as a causal estimate of domain shift alone.

The HEEDB-scale analysis (Table~\ref{tab:heedb_rank1_grid}) supports the same interpretation. Performance declines as gallery size grows and improves as more examinations per patient are available. Thus, larger galleries make identification systematically harder, whereas multiple templates per identity partially compensate for that loss.

\paragraph{Contextualization with respect to previous literature.}
Comparison with previous literature requires caution. Many published ECG identification results are reported on very small cohorts, often below 300 subjects, where inter-subject confusion is intrinsically limited \cite{ibtehaz2022,alduwaile2021,wang2024ecg}. In addition, temporal distance between enrollment and test is often not controlled or not reported. Where it is reported, performance can deteriorate substantially even over weeks or months \cite{chee2022,alduwaile2021}. The present study operates in a more demanding regime: galleries of 63{,}895 and 118{,}756 patients, guaranteed temporal separation in dataset construction, and an explicit stress test up to five years.

The closest large-scale reference is Melzi et al.~\cite{melzi2023}, who report 96.46\% identification accuracy on a proprietary multi-session cohort of about 55{,}967 subjects. The results reported here are numerically lower, but they were obtained under a harder and publicly verifiable setting, with larger galleries and stronger domain heterogeneity. They should therefore be interpreted as results from a severe closed-set benchmark rather than compared directly with small-scale studies.

\paragraph{Temporal Stress Test (MIMIC-TST, HEEDB-TST).}
Table~\ref{tab:tst_const_rank1} shows a monotonic decay of Rank@1 as temporal gap increases from 1 to 5 years in both external datasets, from 0.7853 to 0.6433 on MIMIC and from 0.6864 to 0.5560 on HEEDB. Temporal distance is therefore a major stress factor: identity information remains detectable, but longitudinal drift progressively degrades retrieval. Static test sets are consequently insufficient to characterize longitudinal robustness.

\paragraph{Reranking (HEEDB-RR): gains, limits, and calibration.}
Table~\ref{tab:rr_summary} shows that reranking is not uniformly beneficial. AS-norm provides the best retrieval gains, whereas diffusion and some query-expansion variants can degrade performance. The focused AS-norm analysis (Table~\ref{tab:asnorm_focus_n}) further shows that hyperparameter selection is multi-objective: the highest Rank@1 in the focused sweep is 0.8005 at $N=4$, but calibration and selective-prediction metrics vary across settings. Reranking should therefore be selected jointly on retrieval and confidence criteria rather than on Rank@1 alone.

An important negative result concerns diffusion. In the tested settings, manifold propagation altered the score distribution in a way that was poorly matched to the logistic confidence model, yielding zero high-confidence coverage in the supplementary confidence tables (Table~S25). This indicates that reranking and confidence calibration cannot be treated as independent components.

\paragraph{Informative value of confidence scores.}
Confidence metrics add information beyond Rank@K and TAR@FAR. In all domains, $\bar{c}_{y=1} > \bar{c}_{y=0}$, so correct identifications receive higher confidence than errors. However, calibration and selective prediction deteriorate out of domain: at $\tau = 0.90$, coverage decreases from 88.05\% on ASUGI-DB to 58.89\% on MIMIC-GC and 35.73\% on HEEDB-GC, while residual error increases from 0.14\% to 1.66\% and 3.31\%. Confidence is therefore informative, but clearly domain-dependent.

\paragraph{Limitations and future directions.}
The main limitation is the closed-set protocol. The present results characterize a large-scale leave-one-out benchmark, not an open-set deployment scenario with unknown identities, rejection rules, and operating thresholds. A second limitation is the asymmetric curation between internal and external domains, which prevents the observed performance gap from being attributed solely to domain shift. A third limitation concerns confidence calibration: the calibrator was studied only on GC and HEEDB-RR, and its portability across domains and protocols remains open. Finally, verification analysis was not a primary target of the study; TAR@FAR was reported for comparability, but full ROC-based operating-threshold analysis remains future work.

An important next step is stratification by diagnosis. Both MIMIC and HEEDB provide structured clinical labels that could be used to quantify how conditions such as atrial fibrillation, bundle branch block, or hypertrophy affect retrieval performance, temporal degradation, and confidence calibration. This would improve both operational interpretation and model understanding.

\paragraph{Integrated reading.}
Taken together, the results define a coherent picture: identity information remains measurable out of domain, but degrades systematically with heterogeneity and temporal distance. Reranking can recover part of the loss, whereas confidence remains informative but less favorable on external corpora. The study therefore supports robustness in a large-scale closed-set benchmark, while also delimiting the conditions under which performance deteriorates.

\section{Conclusions}

This study evaluated ECG biometrics under a substantially harder regime than is typical in the literature: galleries up to 118{,}756 identities, external multi-domain validation, and temporal gaps up to five years. The proposed Inception-v1 1D + ArcFace system preserved measurable discriminative power out of domain, reaching Rank@1 = 0.8291 on MIMIC-GC and 0.6884 on HEEDB-GC.

Across the four evaluation protocols, three findings are central. First, performance declines systematically with both gallery size and temporal distance, showing that scale and longitudinal drift are major stress factors. Second, reranking can partially recover retrieval performance, with the focused AS-norm sweep reaching Rank@1 = 0.8005 on HEEDB-RR, but the gain depends on the joint behavior of ranking and calibration. Third, confidence remains informative across domains, with $\Delta = \bar{c}_{y=1} - \bar{c}_{y=0}$ ranging from 0.3731 to 0.4882 in GC, while high-confidence coverage decreases and residual error increases in the external datasets.

Overall, the results show that ECG identity information remains measurable under large-scale closed-set stress conditions, but that its operational quality is strongly conditioned by domain heterogeneity, temporal drift, and the interaction between retrieval and calibration. Open-set validation and deployment-oriented threshold analysis remain future work.

\section*{Data Availability}

The internal clinical data (ASUGI corpus) are owned by ASUGI -- Azienda Sanitaria Universitaria Giuliano Isontina and cannot be publicly distributed due to the confidentiality constraints associated with health data. The MIMIC-IV-ECG \cite{mimic4wdb_010} and HEEDB \cite{koscova2025heedb,heedb_50} corpora are publicly accessible through PhysioNet (\url{https://physionet.org}) and the Brain Data Science Platform (\url{https://bdsp.io}), respectively, upon registration with verified credentials and acceptance of the corresponding Data Use Agreements. All derived datasets (MIMIC-GC, HEEDB-GC, MIMIC-TST, HEEDB-TST, HEEDB-scale, HEEDB-RR) are reproducible by applying the construction pipeline described in Section~\ref{sec:pipeline} to the public corpora.

\section*{Supplementary Materials}

Extended tables, additional confidence analyses, dataset-construction details, and supporting technical material are provided in the companion Supplementary Materials file \texttt{materiali\_aggiuntivi-EN.tex}. All references in the main text to Tables~S1--S31 refer to that file.

\section*{Ethics Statement}

The clinical data of the ASUGI corpus were collected within the routine diagnostic activity of ASUGI -- Azienda Sanitaria Universitaria Giuliano Isontina. Their use for research purposes was approved by the competent Ethics Committee. The data were processed in anonymized form, with linkage to a patient identifier devoid of direct personal information, in compliance with the applicable personal-data protection regulations (EU Regulation 2016/679 -- GDPR). The public MIMIC-IV-ECG and HEEDB corpora were used in compliance with their respective Data Use Agreements; both datasets were de-identified by their curators prior to public release.

\bibliographystyle{plain}
\bibliography{references-EN}

\end{document}


\maketitle

\begin{landscape}
\begin{table}[p]
\centering
\caption{Corpora and derived experimental datasets used in the study.}
\label{tab:corpora_datasets}
\small
\begin{tabular}{llp{3.5cm}rp{9cm}}
\toprule
Corpus & Short name & Protocol & N datasets & Purpose \\
\midrule
ASUGI & ASUGI-DB & Training / Validation / Internal test & 3 & Patient-disjoint splits in three configurations (db1000/1001/1002); db1001 adopted as reference. \\
\midrule
\multirow{2}{*}{MIMIC} & MIMIC-GC & General Comparability & 1 & Closed-set evaluation on the entire eligible pool of the MIMIC corpus (231{,}329 ECGs, 63{,}895 patients). \\
 & MIMIC-TST & Temporal Stress Test & 5 & One pair per patient for each of the 5 temporal targets (1--5 years $\pm$3 months), with constant gallery at $P=14{,}521$. \\
\midrule
\multirow{4}{*}{HEEDB} & HEEDB-GC & General Comparability & 1 & Closed-set evaluation on the entire eligible pool of the HEEDB/i0006 corpus, 500 Hz ECGs only (385{,}079 ECGs, 118{,}756 patients). \\
 & HEEDB-TST & Temporal Stress Test & 5 & One pair per patient for each of the 5 temporal targets (1--5 years $\pm$3 months), with constant gallery at $P=14{,}763$. \\
 & HEEDB-scale & Scale analysis & $N_\text{gallery} \times N_\text{exams}$ & Stratified sample with full grid of gallery size (500--7000, 14 values) $\times$ exams per patient (2--7, 6 values) = 84 configurations. \\
 & HEEDB-RR & Reranking & 1 & Proportional stratified sample from HEEDB-GC ($P=10{,}000$, seed 42, all ECGs of the selected patients). Fixed gallery; chosen for manageable computational cost, statistical stability (moments 1--4 close to full HEEDB), and fast iteration over configuration grids for multiple methods (AS/S/Z/C/T-norm, diffusion, AQE). \\
\bottomrule
\end{tabular}
\end{table}
\end{landscape}

\begin{table}[ht]
\centering
\caption{ASUGIDB split construction for the three training datasets (patient-disjoint).}
\label{tab:internal_train_val_test_splits}
\small
\begin{tabular}{llrrrrr}
\toprule
Dataset ID & Split & ECGs & Patients & Min ECG/pat. & Mean ECG/pat. & Max ECG/pat. \\
\midrule
1000 & Train      & 89{,}576 & 21{,}407 & 3 & 4.18 & 10 \\
1000 & Validation & 37{,}352 & 15{,}836 & 2 & 2.36 & 10 \\
1000 & Test       & 37{,}512 & 15{,}836 & 2 & 2.37 & 10 \\
\midrule
1001 & Train      & 60{,}153 & 11{,}647 & 4 & 5.16 & 10 \\
1001 & Validation & 51{,}988 & 20{,}716 & 2 & 2.51 & 10 \\
1001 & Test       & 52{,}299 & 20{,}716 & 2 & 2.52 & 10 \\
\midrule
1002 & Train      & 39{,}473 &  6{,}479 & 5 & 6.09 & 10 \\
1002 & Validation & 62{,}460 & 23{,}300 & 2 & 2.68 & 10 \\
1002 & Test       & 62{,}507 & 23{,}300 & 2 & 2.68 & 10 \\
\bottomrule
\end{tabular}
\par\medskip\footnotesize\textit{Note: the datasets differ in the minimum number of ECGs per patient required in training (3/4/5, respectively). Validation and test always require at least 2 ECGs per patient. In the present study, db1001 was adopted as the reference configuration.}
\end{table}

\begin{table}[ht]
\centering
\caption{Summary technical specifications of the ELI250 device (Mortara Instrument).}
\label{tab:eli250_specs}
\small
\begin{tabular}{ll}
\toprule
Parameter & Value \\
\midrule
Manufacturer                    & Mortara Instrument (Milwaukee, WI, USA) \\
Leads                           & 12 simultaneous (I, II, III, aVR, aVL, aVF, V1--V6) \\
Sampling frequency              & 1000 Hz (recording); 40{,}000 Hz (pacemaker detection) \\
ADC resolution                  & 20 bit, 1.17 $\mu$V/LSB \\
Frequency response              & 0.05--300 Hz \\
Input impedance                 & $>$50 M$\Omega$ (ANSI/AAMI EC11) \\
CMRR                            & $>$140 dB \\
Available filters               & Baseline, notch 50/60 Hz, low-pass 40/150/300 Hz \\
Interpretation algorithm        & Mortara VERITAS\texttrademark\ (sex- and age-specific criteria) \\
Connectivity                    & Serial, modem, LAN, WLAN (WEP/WPA-PSK/WPA2/PEAP) \\
Internal storage                & 40--60 ECGs (expandable to 150--200) \\
Power supply                    & Mains or battery \\
Supported patients              & Adults and pediatric patients \\
\bottomrule
\end{tabular}
\end{table}

\begin{table}[ht]
\centering
\caption{HEEDB -- Rank@5 as a function of gallery size (500--7000) and number of examinations per patient (2--7). Mean over seeds 0--5.}
\label{tab:heedb_rank5_grid}
\footnotesize
\begin{tabular}{rrrrrrrr}
\toprule
Gallery & Ex=2 & Ex=3 & Ex=4 & Ex=5 & Ex=6 & Ex=7 & Mean \\
\midrule
500  & 0.8107 & 0.9036 & 0.9327 & 0.9360 & 0.9456 & 0.9465 & 0.9125 \\
1000 & 0.7937 & 0.8825 & 0.9186 & 0.9271 & 0.9361 & 0.9387 & 0.8994 \\
1500 & 0.7824 & 0.8747 & 0.9103 & 0.9216 & 0.9294 & 0.9342 & 0.8921 \\
2000 & 0.7731 & 0.8673 & 0.9064 & 0.9170 & 0.9270 & 0.9318 & 0.8871 \\
2500 & 0.7646 & 0.8634 & 0.9017 & 0.9141 & 0.9239 & 0.9290 & 0.8828 \\
3000 & 0.7568 & 0.8581 & 0.8976 & 0.9104 & 0.9211 & 0.9264 & 0.8784 \\
3500 & 0.7503 & 0.8539 & 0.8937 & 0.9084 & 0.9185 & 0.9241 & 0.8748 \\
4000 & 0.7462 & 0.8512 & 0.8903 & 0.9062 & 0.9163 & 0.9217 & 0.8720 \\
4500 & 0.7411 & 0.8487 & 0.8881 & 0.9044 & 0.9138 & 0.9198 & 0.8693 \\
5000 & 0.7384 & 0.8458 & 0.8860 & 0.9025 & 0.9120 & 0.9185 & 0.8672 \\
5500 & 0.7364 & 0.8434 & 0.8836 & 0.9008 & 0.9102 & 0.9171 & 0.8653 \\
6000 & 0.7344 & 0.8411 & 0.8817 & 0.8992 & 0.9085 & 0.9160 & 0.8635 \\
6500 & 0.7329 & 0.8392 & 0.8801 & 0.8979 & 0.9068 & 0.9145 & 0.8619 \\
7000 & 0.7298 & 0.8375 & 0.8780 & 0.8967 & 0.9059 & 0.9133 & 0.8602 \\
\midrule
Mean & 0.7565 & 0.8579 & 0.8963 & 0.9101 & 0.9197 & 0.9251 & 0.8776 \\
\bottomrule
\end{tabular}
\end{table}

\begin{table}[ht]
\centering
\caption{HEEDB -- Rank@10 as a function of gallery size (500--7000) and number of examinations per patient (2--7). Mean over seeds 0--5.}
\label{tab:heedb_rank10_grid}
\footnotesize
\begin{tabular}{rrrrrrrr}
\toprule
Gallery & Ex=2 & Ex=3 & Ex=4 & Ex=5 & Ex=6 & Ex=7 & Mean \\
\midrule
500  & 0.8292 & 0.9173 & 0.9426 & 0.9453 & 0.9542 & 0.9549 & 0.9239 \\
1000 & 0.8158 & 0.8993 & 0.9305 & 0.9369 & 0.9447 & 0.9476 & 0.9125 \\
1500 & 0.8046 & 0.8888 & 0.9229 & 0.9324 & 0.9397 & 0.9434 & 0.9053 \\
2000 & 0.7954 & 0.8833 & 0.9193 & 0.9291 & 0.9374 & 0.9414 & 0.9010 \\
2500 & 0.7862 & 0.8802 & 0.9151 & 0.9259 & 0.9346 & 0.9391 & 0.8969 \\
3000 & 0.7793 & 0.8752 & 0.9115 & 0.9226 & 0.9324 & 0.9367 & 0.8929 \\
3500 & 0.7731 & 0.8710 & 0.9081 & 0.9213 & 0.9303 & 0.9346 & 0.8897 \\
4000 & 0.7701 & 0.8685 & 0.9048 & 0.9189 & 0.9280 & 0.9326 & 0.8871 \\
4500 & 0.7658 & 0.8666 & 0.9027 & 0.9173 & 0.9254 & 0.9310 & 0.8848 \\
5000 & 0.7635 & 0.8639 & 0.9004 & 0.9156 & 0.9239 & 0.9296 & 0.8828 \\
5500 & 0.7615 & 0.8615 & 0.8984 & 0.9139 & 0.9222 & 0.9284 & 0.8810 \\
6000 & 0.7596 & 0.8597 & 0.8967 & 0.9123 & 0.9207 & 0.9274 & 0.8794 \\
6500 & 0.7585 & 0.8579 & 0.8952 & 0.9113 & 0.9194 & 0.9261 & 0.8781 \\
7000 & 0.7553 & 0.8566 & 0.8930 & 0.9099 & 0.9185 & 0.9250 & 0.8764 \\
\midrule
Mean & 0.7799 & 0.8750 & 0.9101 & 0.9223 & 0.9308 & 0.9356 & 0.8923 \\
\bottomrule
\end{tabular}
\end{table}

\begin{table}[ht]
\centering
\caption{HEEDB -- Rank@100 as a function of gallery size (500--7000) and number of examinations per patient (2--7). Mean over seeds 0--5.}
\label{tab:heedb_rank100_grid}
\footnotesize
\begin{tabular}{rrrrrrrr}
\toprule
Gallery & Ex=2 & Ex=3 & Ex=4 & Ex=5 & Ex=6 & Ex=7 & Mean \\
\midrule
500  & 0.8910 & 0.9516 & 0.9650 & 0.9690 & 0.9764 & 0.9750 & 0.9547 \\
1000 & 0.8758 & 0.9385 & 0.9585 & 0.9640 & 0.9699 & 0.9703 & 0.9462 \\
1500 & 0.8674 & 0.9316 & 0.9532 & 0.9598 & 0.9661 & 0.9675 & 0.9409 \\
2000 & 0.8599 & 0.9276 & 0.9505 & 0.9577 & 0.9634 & 0.9656 & 0.9374 \\
2500 & 0.8521 & 0.9253 & 0.9479 & 0.9556 & 0.9611 & 0.9640 & 0.9343 \\
3000 & 0.8459 & 0.9220 & 0.9458 & 0.9534 & 0.9594 & 0.9626 & 0.9315 \\
3500 & 0.8414 & 0.9183 & 0.9440 & 0.9519 & 0.9578 & 0.9611 & 0.9291 \\
4000 & 0.8386 & 0.9165 & 0.9421 & 0.9504 & 0.9563 & 0.9597 & 0.9273 \\
4500 & 0.8354 & 0.9144 & 0.9402 & 0.9495 & 0.9546 & 0.9585 & 0.9255 \\
5000 & 0.8337 & 0.9124 & 0.9387 & 0.9482 & 0.9533 & 0.9575 & 0.9240 \\
5500 & 0.8316 & 0.9110 & 0.9371 & 0.9471 & 0.9521 & 0.9564 & 0.9225 \\
6000 & 0.8302 & 0.9094 & 0.9357 & 0.9462 & 0.9512 & 0.9557 & 0.9214 \\
6500 & 0.8286 & 0.9082 & 0.9345 & 0.9454 & 0.9503 & 0.9549 & 0.9203 \\
7000 & 0.8268 & 0.9071 & 0.9332 & 0.9444 & 0.9495 & 0.9541 & 0.9192 \\
\midrule
Mean & 0.8470 & 0.9210 & 0.9447 & 0.9530 & 0.9587 & 0.9616 & 0.9310 \\
\bottomrule
\end{tabular}
\end{table}

\begin{table}[ht]
\centering
\caption{HEEDB -- Rank@200 as a function of gallery size (500--7000) and number of examinations per patient (2--7). Mean over seeds 0--5.}
\label{tab:heedb_rank200_grid}
\footnotesize
\begin{tabular}{rrrrrrrr}
\toprule
Gallery & Ex=2 & Ex=3 & Ex=4 & Ex=5 & Ex=6 & Ex=7 & Mean \\
\midrule
500  & 0.9128 & 0.9593 & 0.9712 & 0.9753 & 0.9815 & 0.9804 & 0.9634 \\
1000 & 0.8928 & 0.9487 & 0.9652 & 0.9700 & 0.9753 & 0.9757 & 0.9546 \\
1500 & 0.8836 & 0.9420 & 0.9605 & 0.9664 & 0.9720 & 0.9729 & 0.9496 \\
2000 & 0.8778 & 0.9375 & 0.9583 & 0.9647 & 0.9697 & 0.9712 & 0.9465 \\
2500 & 0.8709 & 0.9351 & 0.9558 & 0.9630 & 0.9680 & 0.9697 & 0.9437 \\
3000 & 0.8642 & 0.9329 & 0.9534 & 0.9612 & 0.9661 & 0.9687 & 0.9411 \\
3500 & 0.8592 & 0.9300 & 0.9518 & 0.9598 & 0.9649 & 0.9673 & 0.9388 \\
4000 & 0.8562 & 0.9284 & 0.9502 & 0.9583 & 0.9633 & 0.9661 & 0.9371 \\
4500 & 0.8531 & 0.9263 & 0.9487 & 0.9574 & 0.9616 & 0.9648 & 0.9353 \\
5000 & 0.8516 & 0.9242 & 0.9470 & 0.9561 & 0.9605 & 0.9639 & 0.9339 \\
5500 & 0.8497 & 0.9229 & 0.9458 & 0.9550 & 0.9594 & 0.9631 & 0.9326 \\
6000 & 0.8483 & 0.9215 & 0.9445 & 0.9542 & 0.9586 & 0.9625 & 0.9316 \\
6500 & 0.8468 & 0.9204 & 0.9434 & 0.9535 & 0.9578 & 0.9617 & 0.9306 \\
7000 & 0.8454 & 0.9194 & 0.9421 & 0.9527 & 0.9571 & 0.9610 & 0.9296 \\
\midrule
Mean & 0.8652 & 0.9321 & 0.9527 & 0.9605 & 0.9654 & 0.9678 & 0.9406 \\
\bottomrule
\end{tabular}
\end{table}

\begin{table}[ht]
\centering
\caption{Temporal Stress Test with variable gallery size -- MIMIC (pairs at 1--5 years, $\pm$3 months).}
\label{tab:tst_mimic_var}
\small
\begin{tabular}{crrrrrr}
\toprule
Gap & \#Patients & \#ECGs & Rank@1 & Rank@10 & TAR@1e-3 & TAR@1e-4 \\
\midrule
1y & 32{,}373 & 64{,}746 & 0.7583 & 0.8395 & 0.8546 & 0.7503 \\
2y & 25{,}380 & 50{,}760 & 0.7305 & 0.8196 & 0.8260 & 0.7107 \\
3y & 21{,}021 & 42{,}042 & 0.7010 & 0.7984 & 0.8022 & 0.6724 \\
4y & 17{,}494 & 34{,}988 & 0.6732 & 0.7792 & 0.7751 & 0.6353 \\
5y & 14{,}521 & 29{,}042 & 0.6433 & 0.7537 & 0.7436 & 0.5914 \\
\bottomrule
\end{tabular}
\par\medskip\footnotesize\textit{Note: gallery size shrinks as temporal gap increases, introducing a confounding effect due to gallery size. See the main text for the constant-gallery version.}
\end{table}

\begin{table}[ht]
\centering
\caption{Temporal Stress Test with variable gallery size -- HEEDB (pairs at 1--5 years, $\pm$3 months).}
\label{tab:tst_heedb_var}
\small
\begin{tabular}{crrrrrr}
\toprule
Gap & \#Patients & \#ECGs & Rank@1 & Rank@10 & TAR@1e-3 & TAR@1e-4 \\
\midrule
1y & 53{,}959 & 107{,}918 & 0.6351 & 0.7305 & 0.7586 & 0.6489 \\
2y & 39{,}742 &  79{,}484 & 0.6060 & 0.7078 & 0.7283 & 0.6076 \\
3y & 29{,}484 &  58{,}968 & 0.5784 & 0.6852 & 0.6984 & 0.5625 \\
4y & 21{,}529 &  43{,}058 & 0.5601 & 0.6701 & 0.6738 & 0.5314 \\
5y & 14{,}763 &  29{,}526 & 0.5560 & 0.6693 & 0.6559 & 0.5090 \\
\bottomrule
\end{tabular}
\par\medskip\footnotesize\textit{Note: stronger degradation than on MIMIC, consistent with the larger domain shift. Gallery size shrinks as the temporal gap increases. See the main text for the constant-gallery version.}
\end{table}

\begin{table}[ht]
\centering
\caption{Extended physiological ranges used for outlier removal on ECG fiducial features (ASUGIDB).}
\label{tab:fiducial_ranges_supp}
\small
\begin{tabular}{lrrll}
\toprule
Feature & Min & Max & Unit & Description \\
\midrule
VentricularRate  & 40   & 120  & bpm & Ventricular rate \\
PRInterval       & 100  & 240  & ms  & PR interval duration \\
QRSDuration      & 60   & 150  & ms  & QRS complex duration \\
QTInterval       & 300  & 500  & ms  & QT interval duration \\
QTCorrected      & 300  & 500  & ms  & Corrected QT (Bazett formula) \\
PAxis            & $-$30 & 90  & \textdegree & P-wave axis \\
RAxis            & $-$90 & 120 & \textdegree & R-wave axis \\
TAxis            & $-$30 & 120 & \textdegree & T-wave axis \\
POnset           & 50   & 500  & ms  & P-wave onset \\
POffset          & 150  & 550  & ms  & P-wave offset \\
QOnset           & 200  & 650  & ms  & QRS complex onset \\
QOffset          & 300  & 750  & ms  & QRS complex offset \\
TOffset          & 500  & 1200 & ms  & T-wave offset \\
\bottomrule
\end{tabular}
\par\medskip\footnotesize\textit{Note: these constraints were applied to remove recordings with values plausibly due to noise, artifacts, or fiducial-extraction errors. The ranges are intentionally broader than normal physiological values in order to retain patients with cardiac disease while excluding clearly erroneous measurements.}
\end{table}

\begin{table}[ht]
\centering
\caption{MIMIC-TST with constant gallery ($P=14{,}521$, $\pm$3 months).}
\label{tab:tst_mimic_const_supp}
\small
\begin{tabular}{crrrrr}
\toprule
Gap & Rank@1 & Rank@5 & Rank@10 & TAR@1e-3 & TAR@1e-4 \\
\midrule
1y & 0.7853 & 0.8420 & 0.8608 & 0.8551 & 0.7483 \\
2y & 0.7473 & 0.8112 & 0.8340 & 0.8247 & 0.7097 \\
3y & 0.7157 & 0.7894 & 0.8120 & 0.8055 & 0.6756 \\
4y & 0.6834 & 0.7600 & 0.7873 & 0.7767 & 0.6369 \\
5y & 0.6433 & 0.7263 & 0.7537 & 0.7436 & 0.5914 \\
\bottomrule
\end{tabular}
\par\medskip\footnotesize\textit{Note: at 5y, the results coincide with the variable-gallery test because $P=14{,}521$ is already the natural minimum size.}
\end{table}

\begin{table}[ht]
\centering
\caption{HEEDB-TST with constant gallery ($P=14{,}763$, $\pm$3 months).}
\label{tab:tst_heedb_const_supp}
\small
\begin{tabular}{crrrrr}
\toprule
Gap & Rank@1 & Rank@5 & Rank@10 & TAR@1e-3 & TAR@1e-4 \\
\midrule
1y & 0.6864 & 0.7501 & 0.7725 & 0.7646 & 0.6555 \\
2y & 0.6452 & 0.7164 & 0.7397 & 0.7288 & 0.6053 \\
3y & 0.6047 & 0.6806 & 0.7075 & 0.6946 & 0.5679 \\
4y & 0.5776 & 0.6584 & 0.6873 & 0.6761 & 0.5363 \\
5y & 0.5560 & 0.6382 & 0.6693 & 0.6559 & 0.5090 \\
\bottomrule
\end{tabular}
\par\medskip\footnotesize\textit{Note: at 5y, the results coincide with the variable-gallery test because $P=14{,}763$ is already the natural minimum size.}
\end{table}

\begin{landscape}
\begin{table}[p]
\centering
\caption{Reranking experiment 1 -- HEEDB-RR with external HEEDB cohort (seed=42, variable cohort size).}
\label{tab:rr_ext_heedb}
\footnotesize
\begin{tabular}{rlrrrrr}
\toprule
Cohort size & Method & Rank@1 & Rank@5 & Rank@10 & Rank@100 & Rank@200 \\
\midrule
-- & baseline & 0.7765 & 0.8304 & 0.8471 & 0.8940 & 0.9051 \\
\midrule
100 & asnorm & 0.7767 & 0.8330 & 0.8518 & 0.8980 & 0.9091 \\
100 & cnorm  & 0.7383 & 0.8151 & 0.8385 & 0.8962 & 0.9086 \\
100 & snorm  & 0.7767 & 0.8330 & 0.8518 & 0.8980 & 0.9091 \\
100 & tnorm  & 0.7383 & 0.8151 & 0.8385 & 0.8962 & 0.9086 \\
100 & znorm  & 0.7765 & 0.8304 & 0.8471 & 0.8940 & 0.9051 \\
\midrule
250 & asnorm & 0.7787 & 0.8355 & 0.8526 & 0.8983 & 0.9098 \\
250 & cnorm  & 0.7344 & 0.8171 & 0.8400 & 0.8981 & 0.9093 \\
250 & snorm  & 0.7795 & 0.8346 & 0.8532 & 0.8981 & 0.9095 \\
250 & tnorm  & 0.7503 & 0.8204 & 0.8417 & 0.8976 & 0.9090 \\
250 & znorm  & 0.7765 & 0.8304 & 0.8471 & 0.8940 & 0.9051 \\
\midrule
500 & asnorm & 0.7798 & 0.8375 & 0.8546 & 0.8988 & 0.9097 \\
500 & cnorm  & 0.7427 & 0.8223 & 0.8457 & 0.8980 & 0.9096 \\
500 & snorm  & 0.7807 & 0.8350 & 0.8531 & 0.8981 & 0.9094 \\
500 & tnorm  & 0.7547 & 0.8214 & 0.8436 & 0.8976 & 0.9088 \\
500 & znorm  & 0.7765 & 0.8304 & 0.8471 & 0.8940 & 0.9051 \\
\midrule
1000 & asnorm & 0.7821 & 0.8383 & 0.8561 & 0.8994 & 0.9102 \\
1000 & cnorm  & 0.7490 & 0.8259 & 0.8472 & 0.8997 & 0.9102 \\
1000 & snorm  & 0.7812 & 0.8352 & 0.8544 & 0.8984 & 0.9096 \\
1000 & tnorm  & 0.7564 & 0.8221 & 0.8447 & 0.8979 & 0.9092 \\
1000 & znorm  & 0.7765 & 0.8304 & 0.8471 & 0.8940 & 0.9051 \\
\midrule
2000 & asnorm & 0.7850 & 0.8382 & 0.8559 & 0.8993 & 0.9103 \\
2000 & cnorm  & 0.7535 & 0.8264 & 0.8494 & 0.8998 & 0.9101 \\
2000 & snorm  & 0.7815 & 0.8354 & 0.8538 & 0.8986 & 0.9095 \\
2000 & tnorm  & 0.7556 & 0.8224 & 0.8438 & 0.8977 & 0.9090 \\
2000 & znorm  & 0.7765 & 0.8304 & 0.8471 & 0.8940 & 0.9051 \\
\midrule
3000 & asnorm & 0.7849 & 0.8395 & 0.8569 & 0.8995 & 0.9101 \\
3000 & cnorm  & 0.7583 & 0.8293 & 0.8501 & 0.8993 & 0.9101 \\
3000 & snorm  & 0.7817 & 0.8358 & 0.8537 & 0.8985 & 0.9095 \\
3000 & tnorm  & 0.7566 & 0.8228 & 0.8441 & 0.8979 & 0.9091 \\
3000 & znorm  & 0.7765 & 0.8304 & 0.8471 & 0.8940 & 0.9051 \\
\bottomrule
\end{tabular}
\end{table}
\end{landscape}

\begin{landscape}
\begin{table}[p]
\centering
\caption{Reranking experiment 2 -- HEEDB-RR with external ASUGI-DB cohort (seed=42, variable cohort size).}
\label{tab:rr_ext_internal}
\footnotesize
\begin{tabular}{rlrrrrr}
\toprule
Cohort size & Method & Rank@1 & Rank@5 & Rank@10 & Rank@100 & Rank@200 \\
\midrule
-- & baseline & 0.7765 & 0.8304 & 0.8471 & 0.8940 & 0.9051 \\
\midrule
100 & asnorm & 0.7628 & 0.8235 & 0.8412 & 0.8933 & 0.9052 \\
100 & cnorm  & 0.7106 & 0.7963 & 0.8227 & 0.8875 & 0.9031 \\
100 & snorm  & 0.7628 & 0.8235 & 0.8412 & 0.8933 & 0.9052 \\
100 & tnorm  & 0.7106 & 0.7963 & 0.8227 & 0.8875 & 0.9031 \\
100 & znorm  & 0.7765 & 0.8304 & 0.8471 & 0.8940 & 0.9051 \\
\midrule
250 & asnorm & 0.7535 & 0.8187 & 0.8386 & 0.8914 & 0.9048 \\
250 & cnorm  & 0.6717 & 0.7776 & 0.8090 & 0.8848 & 0.9018 \\
250 & snorm  & 0.7700 & 0.8261 & 0.8440 & 0.8932 & 0.9055 \\
250 & tnorm  & 0.7490 & 0.8146 & 0.8350 & 0.8910 & 0.9038 \\
250 & znorm  & 0.7765 & 0.8304 & 0.8471 & 0.8940 & 0.9051 \\
\midrule
500 & asnorm & 0.7437 & 0.8147 & 0.8351 & 0.8911 & 0.9048 \\
500 & cnorm  & 0.6452 & 0.7646 & 0.7970 & 0.8825 & 0.9014 \\
500 & snorm  & 0.7701 & 0.8265 & 0.8441 & 0.8934 & 0.9054 \\
500 & tnorm  & 0.7511 & 0.8166 & 0.8370 & 0.8912 & 0.9040 \\
500 & znorm  & 0.7765 & 0.8304 & 0.8471 & 0.8940 & 0.9051 \\
\midrule
1000 & asnorm & 0.7482 & 0.8155 & 0.8372 & 0.8911 & 0.9040 \\
1000 & cnorm  & 0.6556 & 0.7696 & 0.8026 & 0.8841 & 0.9012 \\
1000 & snorm  & 0.7721 & 0.8269 & 0.8450 & 0.8936 & 0.9055 \\
1000 & tnorm  & 0.7583 & 0.8203 & 0.8395 & 0.8922 & 0.9047 \\
1000 & znorm  & 0.7765 & 0.8304 & 0.8471 & 0.8940 & 0.9051 \\
\midrule
2000 & asnorm & 0.7468 & 0.8173 & 0.8381 & 0.8906 & 0.9042 \\
2000 & cnorm  & 0.6479 & 0.7705 & 0.8046 & 0.8831 & 0.9009 \\
2000 & snorm  & 0.7735 & 0.8278 & 0.8456 & 0.8938 & 0.9058 \\
2000 & tnorm  & 0.7635 & 0.8230 & 0.8411 & 0.8931 & 0.9051 \\
2000 & znorm  & 0.7765 & 0.8304 & 0.8471 & 0.8940 & 0.9051 \\
\midrule
3000 & asnorm & 0.7472 & 0.8168 & 0.8379 & 0.8904 & 0.9041 \\
3000 & cnorm  & 0.6508 & 0.7680 & 0.8020 & 0.8828 & 0.9011 \\
3000 & snorm  & 0.7740 & 0.8285 & 0.8463 & 0.8938 & 0.9059 \\
3000 & tnorm  & 0.7644 & 0.8238 & 0.8416 & 0.8929 & 0.9052 \\
3000 & znorm  & 0.7765 & 0.8304 & 0.8471 & 0.8940 & 0.9051 \\
\bottomrule
\end{tabular}
\end{table}
\end{landscape}

\begin{table}[ht]
\centering
\caption{Full diffusion-reranking and AQE/$\alpha$QE grid on HEEDB-RR (internal cohort, run \texttt{20260208\_123000\_db1001}).}
\label{tab:rr_diffusion_aqe_grid}
\small
\begin{tabular}{llrrrrr}
\toprule
Algorithm & Configuration & Rank@1 & Rank@5 & Rank@10 & Rank@100 & Rank@200 \\
\midrule
Baseline  & -- & 0.7765 & 0.8304 & 0.8471 & 0.8940 & 0.9051 \\
\midrule
Diffusion & K200\_lk10\_$\alpha$0.950\_it5  & 0.7659 & 0.8241 & 0.8430 & 0.8928 & 0.9051 \\
Diffusion & K200\_lk15\_$\alpha$0.950\_it8  & 0.7488 & 0.8164 & 0.8372 & 0.8914 & 0.9051 \\
Diffusion & K300\_lk20\_$\alpha$0.970\_it10 & 0.7488 & 0.8160 & 0.8373 & 0.8913 & 0.9033 \\
Diffusion & K400\_lk30\_$\alpha$0.990\_it20 & 0.6894 & 0.7908 & 0.8197 & 0.8864 & 0.9009 \\
\midrule
AQE/$\alpha$QE & K3,  $\alpha$=1.0 & 0.7358 & 0.8148 & 0.8210 & 0.8582 & 0.8738 \\
AQE/$\alpha$QE & K3,  $\alpha$=2.0 & 0.7534 & 0.8157 & 0.8228 & 0.8670 & 0.8826 \\
AQE/$\alpha$QE & K3,  $\alpha$=3.0 & 0.7626 & 0.8174 & 0.8259 & 0.8750 & 0.8907 \\
AQE/$\alpha$QE & K5,  $\alpha$=1.0 & 0.6830 & 0.8016 & 0.8263 & 0.8565 & 0.8703 \\
AQE/$\alpha$QE & K5,  $\alpha$=2.0 & 0.7189 & 0.8107 & 0.8287 & 0.8626 & 0.8770 \\
AQE/$\alpha$QE & K5,  $\alpha$=3.0 & 0.7404 & 0.8157 & 0.8310 & 0.8695 & 0.8845 \\
AQE/$\alpha$QE & K10, $\alpha$=1.0 & 0.6080 & 0.7288 & 0.7885 & 0.8560 & 0.8688 \\
AQE/$\alpha$QE & K10, $\alpha$=2.0 & 0.6627 & 0.7562 & 0.8050 & 0.8598 & 0.8731 \\
AQE/$\alpha$QE & K10, $\alpha$=3.0 & 0.7007 & 0.7778 & 0.8165 & 0.8657 & 0.8786 \\
AQE/$\alpha$QE & K20, $\alpha$=1.0 & 0.5381 & 0.6639 & 0.7235 & 0.8411 & 0.8600 \\
AQE/$\alpha$QE & K20, $\alpha$=2.0 & 0.6008 & 0.7033 & 0.7518 & 0.8495 & 0.8662 \\
AQE/$\alpha$QE & K20, $\alpha$=3.0 & 0.6530 & 0.7376 & 0.7774 & 0.8569 & 0.8725 \\
\bottomrule
\end{tabular}
\par\medskip\footnotesize\textit{Note: see Table~\ref{tab:rr_config_legend} for the legend of configuration codes.}
\end{table}

\begin{table}[ht]
\centering
\caption{Legend of the configuration codes used in the reranking tables.}
\label{tab:rr_config_legend}
\small
\begin{tabular}{llp{8cm}}
\toprule
Format & Parameters & Description \\
\midrule
\texttt{K400\_C500\_S42} & K, C, S & $K$ = shortlist size; $C$ = cohort size; $S$ = seed. Used for S-norm, Z-norm, and C-norm. \\
\texttt{K400\_N200\_scan2000} & K, N, scan & $K$ = shortlist size; $N$ = number of top-cohort samples used for adaptive statistics; \textit{scan} = candidates scanned to estimate the statistics. Used for AS-norm. \\
\texttt{K3\_a2.0} & K, $\alpha$ & $K$ = number of neighbors for query expansion; $\alpha$ = QE weight exponent. Used for AQE/$\alpha$QE. \\
\texttt{K200\_lk10\_a0p950\_it5} & K, lk, $\alpha$, it & $K$ = shortlist; \textit{lk} = local\_k of the k-NN graph; \textit{a0p950} = $\alpha=0.950$; \textit{it} = diffusion iterations. Used for diffusion. \\
\texttt{baseline} / \texttt{--} & -- & No reranking; original ranking from cosine embeddings. \\
\bottomrule
\end{tabular}
\end{table}

\begin{table}[ht]
\centering
\caption{Reranking experiment 3 -- HEEDB-RR internal reranking (internal cohort, no external cohort).}
\label{tab:rr_internal}
\small
\begin{tabular}{llrrrrr}
\toprule
Method & Configuration & Rank@1 & Rank@5 & Rank@10 & Rank@100 & Rank@200 \\
\midrule
baseline  & --                             & 0.7765 & 0.8304 & 0.8471 & 0.8940 & 0.9051 \\
\midrule
asnorm    & K400\_N200\_scan2000           & 0.7931 & 0.8417 & 0.8573 & 0.8994 & 0.9099 \\
snorm     & K400\_C500\_S42                & 0.7766 & 0.8317 & 0.8490 & 0.8960 & 0.9071 \\
znorm     & K400\_C500\_S42                & 0.7485 & 0.8168 & 0.8384 & 0.8931 & 0.9046 \\
cnorm     & K400\_C500\_S42                & 0.7489 & 0.8175 & 0.8395 & 0.8932 & 0.9045 \\
\midrule
diffusion & K200\_lk10\_$\alpha$0.950\_it5  & 0.7659 & 0.8241 & 0.8430 & 0.8928 & 0.9051 \\
diffusion & K200\_lk15\_$\alpha$0.950\_it8  & 0.7488 & 0.8164 & 0.8372 & 0.8914 & 0.9051 \\
diffusion & K300\_lk20\_$\alpha$0.970\_it10 & 0.7488 & 0.8160 & 0.8373 & 0.8913 & 0.9033 \\
diffusion & K400\_lk30\_$\alpha$0.990\_it20 & 0.6894 & 0.7908 & 0.8197 & 0.8864 & 0.9009 \\
\midrule
aqe & K3, $\alpha$=1.0  & 0.7358 & 0.8148 & 0.8210 & 0.8582 & 0.8738 \\
aqe & K3, $\alpha$=2.0  & 0.7534 & 0.8157 & 0.8228 & 0.8670 & 0.8826 \\
aqe & K3, $\alpha$=3.0  & 0.7626 & 0.8174 & 0.8259 & 0.8750 & 0.8907 \\
aqe & K5, $\alpha$=1.0  & 0.6830 & 0.8016 & 0.8263 & 0.8565 & 0.8703 \\
aqe & K5, $\alpha$=2.0  & 0.7189 & 0.8107 & 0.8287 & 0.8626 & 0.8770 \\
aqe & K5, $\alpha$=3.0  & 0.7404 & 0.8157 & 0.8310 & 0.8695 & 0.8845 \\
aqe & K10, $\alpha$=1.0 & 0.6080 & 0.7288 & 0.7885 & 0.8560 & 0.8688 \\
aqe & K10, $\alpha$=2.0 & 0.6627 & 0.7562 & 0.8050 & 0.8598 & 0.8731 \\
aqe & K10, $\alpha$=3.0 & 0.7007 & 0.7778 & 0.8165 & 0.8657 & 0.8786 \\
aqe & K20, $\alpha$=1.0 & 0.5381 & 0.6639 & 0.7235 & 0.8411 & 0.8600 \\
aqe & K20, $\alpha$=2.0 & 0.6008 & 0.7033 & 0.7518 & 0.8495 & 0.8662 \\
aqe & K20, $\alpha$=3.0 & 0.6530 & 0.7376 & 0.7774 & 0.8569 & 0.8725 \\
\bottomrule
\end{tabular}
\par\medskip\footnotesize\textit{Note: run \texttt{20260208\_123000\_db1001}, best epoch 99.}
\end{table}

\begin{landscape}
\begin{table}[p]
\centering
\caption{Best per algorithm: best observed configuration for each reranking method on HEEDB-RR.}
\label{tab:rr_best_per_algorithm}
\scriptsize
\begin{tabular}{llrrrrrrrr}
\toprule
Algorithm & Config & R@1 & R@5 & R@10 & Acc@0.5 & Brier & ECE & Cov@0.90 & Err@0.90 \\
\midrule
baseline  & --                                       & 0.7765 & 0.8304 & 0.8471 & --     & --     & --     & --     & -- \\
asnorm    & internal\_K400\_N200\_scan2000           & 0.7931 & 0.8417 & 0.8573 & 0.9216 & 0.0572 & 0.0319 & 0.6429 & 0.0031 \\
snorm     & ext\_heedb\_size3000\_K400\_N100\_seed42 & 0.7817 & 0.8358 & 0.8537 & 0.8633 & 0.1038 & 0.0349 & 0.4614 & 0.0362 \\
znorm     & ext\_heedb\_size100\_K400\_N100\_seed42  & 0.7765 & 0.8304 & 0.8471 & 0.8508 & 0.1122 & 0.0492 & 0.4073 & 0.0411 \\
tnorm     & ext\_internal\_size3000\_K400\_N100\_seed42 & 0.7644 & 0.8238 & 0.8416 & 0.8538 & 0.1039 & 0.0377 & 0.4817 & 0.0327 \\
alphaqe   & K3\_a3.0                                 & 0.7626 & 0.8174 & 0.8259 & 0.8609 & 0.0975 & 0.0439 & 0.5128 & 0.0201 \\
cnorm     & ext\_heedb\_size3000\_K400\_N100\_seed42 & 0.7583 & 0.8293 & 0.8501 & 0.8794 & 0.0866 & 0.0290 & 0.5309 & 0.0148 \\
diffusion & K200\_lk15\_a0.950\_it8                  & 0.2567 & 0.4528 & 0.5443 & 0.8568 & 0.1107 & 0.0655 & 0.0000 & NaN \\
\bottomrule
\end{tabular}
\par\medskip\footnotesize\textit{Note: for each algorithm, the best configuration selected from the result files is reported. R@k = Rank@k; Acc@0.5 = binary decision accuracy at threshold 0.5; Brier and ECE measure calibration quality; Cov@0.90 and Err@0.90 denote coverage and error, respectively, when only predictions with confidence at least 0.90 are accepted.}
\end{table}
\end{landscape}

\begin{landscape}
\begin{table}[p]
\centering
\caption{AS-norm focus as a function of $N$ (HEEDB-RR, no external cohort): full table with retrieval and confidence metrics.}
\label{tab:asnorm_focus_full}
\scriptsize
\begin{tabular}{rrrrrrrrrrrr}
\toprule
$N$ & Rank@1 & Rank@5 & Rank@10 & Acc@0.5 & Brier & ECE & Conf mean & Conf mean y1 & Conf mean y0 & Cov@0.90 & Err@0.90 \\
\midrule
1  & 0.7943 & 0.8387 & 0.8552 & 0.9623 & 0.0391 & 0.0690 & 0.7862 & 0.9291 & 0.2372 & 0.6573 & 0.0000 \\
2  & 0.7977 & 0.8326 & 0.8472 & 0.7974 & 0.1580 & 0.0332 & 0.7990 & 0.8017 & 0.7882 & 0.0182 & 0.0000 \\
3  & 0.8003 & 0.8371 & 0.8527 & 0.9499 & 0.0653 & 0.1403 & 0.7907 & 0.8847 & 0.4152 & 0.5124 & 0.0000 \\
4  & 0.8005 & 0.8386 & 0.8541 & 0.9682 & 0.0536 & 0.1051 & 0.7908 & 0.9017 & 0.3479 & 0.5710 & 0.0000 \\
5  & 0.7999 & 0.8389 & 0.8543 & 0.9653 & 0.0482 & 0.0970 & 0.7904 & 0.9106 & 0.3123 & 0.6009 & 0.0000 \\
6  & 0.7995 & 0.8393 & 0.8549 & 0.9648 & 0.0460 & 0.0885 & 0.7903 & 0.9148 & 0.2962 & 0.6140 & 0.0000 \\
7  & 0.7996 & 0.8398 & 0.8552 & 0.9078 & 0.0840 & 0.0592 & 0.7962 & 0.8805 & 0.4598 & 0.4151 & 0.0000 \\
8  & 0.7987 & 0.8402 & 0.8556 & 0.9071 & 0.0837 & 0.0570 & 0.7953 & 0.8812 & 0.4543 & 0.4177 & 0.0000 \\
9  & 0.7988 & 0.8405 & 0.8560 & 0.9633 & 0.0443 & 0.0772 & 0.7897 & 0.9190 & 0.2789 & 0.6293 & 0.0000 \\
10 & 0.7987 & 0.8407 & 0.8561 & 0.9626 & 0.0443 & 0.0744 & 0.7897 & 0.9195 & 0.2769 & 0.6319 & 0.0000 \\
\bottomrule
\end{tabular}
\par\medskip\footnotesize\textit{Note: full table for the AS-norm focus reported in the manuscript. Conf mean = overall mean confidence; Conf mean y1 = mean confidence on correct identifications; Conf mean y0 = mean confidence on incorrect identifications.}
\end{table}
\end{landscape}

\section*{Reranking tables with confidence}

Baseline Rank@1 = 0.7765. Run \texttt{20260208\_123000\_db1001}, HEEDB-RR dataset ($P=10{,}000$, seed~42).
Metrics: \textbf{R@1} = reranked Rank@1; \textbf{dR@1} = delta vs baseline; \textbf{Acc} = accuracy@0.5; \textbf{Brier} = Brier score; \textbf{ECE} = expected calibration error; \textbf{$\mu_{y=1}$} / \textbf{$\mu_{y=0}$} = mean confidence on correct/incorrect samples; \textbf{Cov90} / \textbf{Err90} = coverage and error at 90\% threshold.

\begin{landscape}
\begin{table}[p]
\centering
\caption{Reranking with confidence -- AS-norm (13 configurations).}
\label{tab:conf_asnorm}
\footnotesize
\begin{tabular}{lrrrrrrrrrr}
\toprule
Config & R@1 & dR@1 & Acc & Brier & ECE & $\mu_{y=1}$ & $\mu_{y=0}$ & Cov90 & Err90 \\
\midrule
internal\_K400\_N200\_scan2000            & 0.7931 & +0.0166 & 0.9225 & 0.0572 & 0.0257 & 0.9139 & 0.2883 & 0.6355 & 0.0031 \\
ext\_heedb\_size100\_K400\_N100\_seed42   & 0.7767 & +0.0002 & 0.8616 & 0.1087 & 0.0536 & 0.8510 & 0.5287 & 0.4182 & 0.0417 \\
ext\_heedb\_size250\_K400\_N100\_seed42   & 0.7787 & +0.0022 & 0.8656 & 0.1009 & 0.0390 & 0.8582 & 0.4901 & 0.4571 & 0.0296 \\
ext\_heedb\_size500\_K400\_N100\_seed42   & 0.7798 & +0.0033 & 0.8810 & 0.0868 & 0.0276 & 0.8801 & 0.4220 & 0.5324 & 0.0210 \\
ext\_heedb\_size1000\_K400\_N100\_seed42  & 0.7821 & +0.0056 & 0.8884 & 0.0804 & 0.0166 & 0.8901 & 0.3825 & 0.5735 & 0.0154 \\
ext\_heedb\_size2000\_K400\_N100\_seed42  & 0.7850 & +0.0085 & 0.8903 & 0.0789 & 0.0126 & 0.8931 & 0.3706 & 0.5922 & 0.0139 \\
ext\_heedb\_size3000\_K400\_N100\_seed42  & 0.7849 & +0.0084 & 0.8946 & 0.0759 & 0.0229 & 0.8953 & 0.3673 & 0.5898 & 0.0114 \\
ext\_internal\_size100\_K400\_N100\_seed42  & 0.7628 & $-$0.0137 & 0.8469 & 0.1104 & 0.0401 & 0.8463 & 0.4991 & 0.4147 & 0.0351 \\
ext\_internal\_size250\_K400\_N100\_seed42  & 0.7535 & $-$0.0230 & 0.8353 & 0.1184 & 0.0369 & 0.8329 & 0.5065 & 0.3756 & 0.0423 \\
ext\_internal\_size500\_K400\_N100\_seed42  & 0.7437 & $-$0.0328 & 0.8243 & 0.1301 & 0.0491 & 0.8153 & 0.5394 & 0.3096 & 0.0527 \\
ext\_internal\_size1000\_K400\_N100\_seed42 & 0.7482 & $-$0.0283 & 0.8226 & 0.1275 & 0.0480 & 0.8225 & 0.5385 & 0.3330 & 0.0494 \\
ext\_internal\_size2000\_K400\_N100\_seed42 & 0.7468 & $-$0.0297 & 0.8262 & 0.1271 & 0.0527 & 0.8198 & 0.5363 & 0.3193 & 0.0481 \\
ext\_internal\_size3000\_K400\_N100\_seed42 & 0.7472 & $-$0.0293 & 0.8210 & 0.1303 & 0.0514 & 0.8152 & 0.5445 & 0.3099 & 0.0510 \\
\bottomrule
\end{tabular}
\end{table}

\begin{table}[p]
\centering
\caption{Reranking with confidence -- S-norm (13 configurations).}
\label{tab:conf_snorm}
\footnotesize
\begin{tabular}{lrrrrrrrrrr}
\toprule
Config & R@1 & dR@1 & Acc & Brier & ECE & $\mu_{y=1}$ & $\mu_{y=0}$ & Cov90 & Err90 \\
\midrule
internal\_K400\_C500\_S42                 & 0.7766 & +0.0001 & 0.8604 & 0.1061 & 0.0373 & 0.8545 & 0.5069 & 0.4420 & 0.0354 \\
ext\_heedb\_size100\_K400\_N100\_seed42   & 0.7767 & +0.0002 & 0.8616 & 0.1087 & 0.0536 & 0.8510 & 0.5287 & 0.4182 & 0.0417 \\
ext\_heedb\_size250\_K400\_N100\_seed42   & 0.7795 & +0.0030 & 0.8622 & 0.1074 & 0.0486 & 0.8535 & 0.5278 & 0.4270 & 0.0430 \\
ext\_heedb\_size500\_K400\_N100\_seed42   & 0.7807 & +0.0042 & 0.8624 & 0.1065 & 0.0419 & 0.8551 & 0.5223 & 0.4364 & 0.0383 \\
ext\_heedb\_size1000\_K400\_N100\_seed42  & 0.7812 & +0.0047 & 0.8640 & 0.1042 & 0.0357 & 0.8604 & 0.5116 & 0.4573 & 0.0374 \\
ext\_heedb\_size2000\_K400\_N100\_seed42  & 0.7815 & +0.0050 & 0.8633 & 0.1042 & 0.0355 & 0.8586 & 0.5099 & 0.4544 & 0.0359 \\
ext\_heedb\_size3000\_K400\_N100\_seed42  & 0.7817 & +0.0052 & 0.8644 & 0.1033 & 0.0354 & 0.8598 & 0.5064 & 0.4594 & 0.0369 \\
ext\_internal\_size100\_K400\_N100\_seed42  & 0.7628 & $-$0.0137 & 0.8469 & 0.1104 & 0.0401 & 0.8463 & 0.4991 & 0.4147 & 0.0351 \\
ext\_internal\_size250\_K400\_N100\_seed42  & 0.7700 & $-$0.0065 & 0.8598 & 0.0998 & 0.0362 & 0.8598 & 0.4661 & 0.4658 & 0.0267 \\
ext\_internal\_size500\_K400\_N100\_seed42  & 0.7701 & $-$0.0064 & 0.8592 & 0.1016 & 0.0361 & 0.8574 & 0.4730 & 0.4550 & 0.0304 \\
ext\_internal\_size1000\_K400\_N100\_seed42 & 0.7721 & $-$0.0044 & 0.8630 & 0.0990 & 0.0361 & 0.8610 & 0.4669 & 0.4688 & 0.0284 \\
ext\_internal\_size2000\_K400\_N100\_seed42 & 0.7735 & $-$0.0030 & 0.8646 & 0.0983 & 0.0342 & 0.8614 & 0.4619 & 0.4770 & 0.0283 \\
ext\_internal\_size3000\_K400\_N100\_seed42 & 0.7740 & $-$0.0025 & 0.8647 & 0.0978 & 0.0340 & 0.8621 & 0.4595 & 0.4805 & 0.0275 \\
\bottomrule
\end{tabular}
\end{table}

\begin{table}[p]
\centering
\caption{Reranking with confidence -- Z-norm (13 configurations). Rank@1 is invariant across all external configurations.}
\label{tab:conf_znorm}
\footnotesize
\begin{tabular}{lrrrrrrrrrr}
\toprule
Config & R@1 & dR@1 & Acc & Brier & ECE & $\mu_{y=1}$ & $\mu_{y=0}$ & Cov90 & Err90 \\
\midrule
internal\_K400\_C500\_S42                 & 0.7485 & $-$0.0280 & 0.8380 & 0.1220 & 0.0395 & 0.8332 & 0.5191 & 0.3724 & 0.0432 \\
ext\_heedb\_size100\_K400\_N100\_seed42   & 0.7765 & 0.0000 & 0.8537 & 0.1111 & 0.0513 & 0.8443 & 0.5344 & 0.3973 & 0.0406 \\
ext\_heedb\_size250\_K400\_N100\_seed42   & 0.7765 & 0.0000 & 0.8568 & 0.1101 & 0.0504 & 0.8455 & 0.5302 & 0.3999 & 0.0402 \\
ext\_heedb\_size500\_K400\_N100\_seed42   & 0.7765 & 0.0000 & 0.8545 & 0.1100 & 0.0412 & 0.8464 & 0.5248 & 0.4109 & 0.0364 \\
ext\_heedb\_size1000\_K400\_N100\_seed42  & 0.7765 & 0.0000 & 0.8581 & 0.1081 & 0.0387 & 0.8492 & 0.5150 & 0.4228 & 0.0363 \\
ext\_heedb\_size2000\_K400\_N100\_seed42  & 0.7765 & 0.0000 & 0.8556 & 0.1086 & 0.0374 & 0.8488 & 0.5172 & 0.4212 & 0.0335 \\
ext\_heedb\_size3000\_K400\_N100\_seed42  & 0.7765 & 0.0000 & 0.8576 & 0.1078 & 0.0375 & 0.8497 & 0.5145 & 0.4231 & 0.0327 \\
ext\_internal\_size100\_K400\_N100\_seed42  & 0.7765 & 0.0000 & 0.8521 & 0.1059 & 0.0333 & 0.8542 & 0.5042 & 0.4384 & 0.0302 \\
ext\_internal\_size250\_K400\_N100\_seed42  & 0.7765 & 0.0000 & 0.8640 & 0.0983 & 0.0331 & 0.8623 & 0.4672 & 0.4781 & 0.0262 \\
ext\_internal\_size500\_K400\_N100\_seed42  & 0.7765 & 0.0000 & 0.8640 & 0.0997 & 0.0339 & 0.8603 & 0.4755 & 0.4662 & 0.0294 \\
ext\_internal\_size1000\_K400\_N100\_seed42 & 0.7765 & 0.0000 & 0.8654 & 0.0977 & 0.0320 & 0.8631 & 0.4626 & 0.4828 & 0.0269 \\
ext\_internal\_size2000\_K400\_N100\_seed42 & 0.7765 & 0.0000 & 0.8682 & 0.0968 & 0.0316 & 0.8638 & 0.4583 & 0.4874 & 0.0275 \\
ext\_internal\_size3000\_K400\_N100\_seed42 & 0.7765 & 0.0000 & 0.8689 & 0.0965 & 0.0313 & 0.8641 & 0.4564 & 0.4887 & 0.0269 \\
\bottomrule
\end{tabular}
\end{table}

\begin{table}[p]
\centering
\caption{Reranking with confidence -- C-norm (13 configurations).}
\label{tab:conf_cnorm}
\footnotesize
\begin{tabular}{lrrrrrrrrrr}
\toprule
Config & R@1 & dR@1 & Acc & Brier & ECE & $\mu_{y=1}$ & $\mu_{y=0}$ & Cov90 & Err90 \\
\midrule
internal\_K400\_C500\_S42                 & 0.7489 & $-$0.0276 & 0.8383 & 0.1196 & 0.0370 & 0.8376 & 0.5089 & 0.3930 & 0.0429 \\
ext\_heedb\_size100\_K400\_N100\_seed42   & 0.7383 & $-$0.0382 & 0.8263 & 0.1297 & 0.0497 & 0.8219 & 0.5297 & 0.3374 & 0.0570 \\
ext\_heedb\_size250\_K400\_N100\_seed42   & 0.7344 & $-$0.0421 & 0.8319 & 0.1235 & 0.0415 & 0.8259 & 0.4956 & 0.3641 & 0.0445 \\
ext\_heedb\_size500\_K400\_N100\_seed42   & 0.7427 & $-$0.0338 & 0.8500 & 0.1074 & 0.0237 & 0.8489 & 0.4328 & 0.4476 & 0.0292 \\
ext\_heedb\_size1000\_K400\_N100\_seed42  & 0.7490 & $-$0.0275 & 0.8662 & 0.0960 & 0.0153 & 0.8690 & 0.3944 & 0.5103 & 0.0221 \\
ext\_heedb\_size2000\_K400\_N100\_seed42  & 0.7535 & $-$0.0230 & 0.8732 & 0.0903 & 0.0113 & 0.8754 & 0.3691 & 0.5334 & 0.0183 \\
ext\_heedb\_size3000\_K400\_N100\_seed42  & 0.7583 & $-$0.0182 & 0.8792 & 0.0869 & 0.0172 & 0.8808 & 0.3727 & 0.5417 & 0.0155 \\
ext\_internal\_size100\_K400\_N100\_seed42  & 0.7106 & $-$0.0659 & 0.8108 & 0.1360 & 0.0452 & 0.7969 & 0.4927 & 0.2834 & 0.0521 \\
ext\_internal\_size250\_K400\_N100\_seed42  & 0.6717 & $-$0.1048 & 0.7788 & 0.1560 & 0.0398 & 0.7615 & 0.4930 & 0.2134 & 0.0640 \\
ext\_internal\_size500\_K400\_N100\_seed42  & 0.6452 & $-$0.1313 & 0.7518 & 0.1732 & 0.0512 & 0.7244 & 0.5096 & 0.1440 & 0.0734 \\
ext\_internal\_size1000\_K400\_N100\_seed42 & 0.6556 & $-$0.1209 & 0.7649 & 0.1677 & 0.0540 & 0.7356 & 0.5074 & 0.1609 & 0.0755 \\
ext\_internal\_size2000\_K400\_N100\_seed42 & 0.6479 & $-$0.1286 & 0.7544 & 0.1732 & 0.0506 & 0.7307 & 0.5167 & 0.1483 & 0.0764 \\
ext\_internal\_size3000\_K400\_N100\_seed42 & 0.6508 & $-$0.1257 & 0.7566 & 0.1708 & 0.0486 & 0.7320 & 0.5087 & 0.1624 & 0.0724 \\
\bottomrule
\end{tabular}
\end{table}

\begin{table}[p]
\centering
\caption{Reranking with confidence -- T-norm (12 configurations, external cohort only).}
\label{tab:conf_tnorm}
\footnotesize
\begin{tabular}{lrrrrrrrrrr}
\toprule
Config & R@1 & dR@1 & Acc & Brier & ECE & $\mu_{y=1}$ & $\mu_{y=0}$ & Cov90 & Err90 \\
\midrule
ext\_heedb\_size100\_K400\_N100\_seed42   & 0.7383 & $-$0.0382 & 0.8263 & 0.1297 & 0.0497 & 0.8219 & 0.5297 & 0.3374 & 0.0570 \\
ext\_heedb\_size250\_K400\_N100\_seed42   & 0.7503 & $-$0.0262 & 0.8361 & 0.1236 & 0.0484 & 0.8317 & 0.5300 & 0.3655 & 0.0537 \\
ext\_heedb\_size500\_K400\_N100\_seed42   & 0.7547 & $-$0.0218 & 0.8399 & 0.1192 & 0.0416 & 0.8374 & 0.5183 & 0.3886 & 0.0446 \\
ext\_heedb\_size1000\_K400\_N100\_seed42  & 0.7564 & $-$0.0200 & 0.8430 & 0.1165 & 0.0398 & 0.8429 & 0.5119 & 0.4066 & 0.0429 \\
ext\_heedb\_size2000\_K400\_N100\_seed42  & 0.7556 & $-$0.0208 & 0.8432 & 0.1162 & 0.0338 & 0.8410 & 0.5046 & 0.4059 & 0.0413 \\
ext\_heedb\_size3000\_K400\_N100\_seed42  & 0.7566 & $-$0.0199 & 0.8448 & 0.1159 & 0.0362 & 0.8415 & 0.5073 & 0.4037 & 0.0413 \\
ext\_internal\_size100\_K400\_N100\_seed42  & 0.7106 & $-$0.0659 & 0.8108 & 0.1360 & 0.0452 & 0.7969 & 0.4927 & 0.2834 & 0.0521 \\
ext\_internal\_size250\_K400\_N100\_seed42  & 0.7490 & $-$0.0275 & 0.8481 & 0.1099 & 0.0331 & 0.8431 & 0.4632 & 0.4169 & 0.0312 \\
ext\_internal\_size500\_K400\_N100\_seed42  & 0.7511 & $-$0.0254 & 0.8499 & 0.1093 & 0.0384 & 0.8451 & 0.4714 & 0.4116 & 0.0359 \\
ext\_internal\_size1000\_K400\_N100\_seed42 & 0.7583 & $-$0.0182 & 0.8540 & 0.1044 & 0.0358 & 0.8536 & 0.4662 & 0.4402 & 0.0314 \\
ext\_internal\_size2000\_K400\_N100\_seed42 & 0.7635 & $-$0.0130 & 0.8551 & 0.1034 & 0.0344 & 0.8554 & 0.4661 & 0.4547 & 0.0314 \\
ext\_internal\_size3000\_K400\_N100\_seed42 & 0.7644 & $-$0.0121 & 0.8570 & 0.1022 & 0.0345 & 0.8562 & 0.4633 & 0.4572 & 0.0296 \\
\bottomrule
\end{tabular}
\end{table}
\end{landscape}

\begin{table}[ht]
\centering
\caption{Reranking with confidence -- Diffusion (4 configurations). \textbf{Note}: anomalous R@1 ($\approx$0.2) due to incompatibility between post-diffusion scores and the logistic-regression calibrator trained on cosine scores.}
\label{tab:conf_diffusion}
\small
\begin{tabular}{lrrrrrrrr}
\toprule
Config & R@1 & dR@1 & Acc & Brier & ECE & $\mu_{y=1}$ & $\mu_{y=0}$ & Cov90 \\
\midrule
K200\_lk10\_$\alpha$0.950\_it5  & 0.2391 & $-$0.5374 & 0.8609 & 0.1069 & 0.0533 & 0.5349 & 0.1471 & 0.0000 \\
K200\_lk15\_$\alpha$0.950\_it8  & 0.2567 & $-$0.5198 & 0.8571 & 0.1105 & 0.0566 & 0.5516 & 0.1543 & 0.0000 \\
K300\_lk20\_$\alpha$0.970\_it10 & 0.2097 & $-$0.5668 & 0.8775 & 0.0963 & 0.0503 & 0.5168 & 0.1291 & 0.0000 \\
K400\_lk30\_$\alpha$0.990\_it20 & 0.1895 & $-$0.5870 & 0.8854 & 0.0911 & 0.0502 & 0.4964 & 0.1173 & 0.0000 \\
\bottomrule
\end{tabular}
\end{table}

\begin{table}[ht]
\centering
\caption{Reranking with confidence -- $\alpha$QE / AQE (12 configurations).}
\label{tab:conf_alphaqe}
\small
\begin{tabular}{lrrrrrrrrrr}
\toprule
Config & R@1 & dR@1 & Acc & Brier & ECE & $\mu_{y=1}$ & $\mu_{y=0}$ & Cov90 & Err90 \\
\midrule
K3, $\alpha$=1.0  & 0.7358 & $-$0.0407 & 0.8323 & 0.1163 & 0.0423 & 0.8366 & 0.4585 & 0.3896 & 0.0247 \\
K3, $\alpha$=2.0  & 0.7534 & $-$0.0231 & 0.8536 & 0.1022 & 0.0439 & 0.8571 & 0.4319 & 0.4660 & 0.0210 \\
K3, $\alpha$=3.0  & 0.7626 & $-$0.0139 & 0.8644 & 0.0962 & 0.0412 & 0.8650 & 0.4195 & 0.5005 & 0.0190 \\
K5, $\alpha$=1.0  & 0.6830 & $-$0.0935 & 0.8005 & 0.1371 & 0.0376 & 0.7962 & 0.4479 & 0.2683 & 0.0303 \\
K5, $\alpha$=2.0  & 0.7189 & $-$0.0576 & 0.8327 & 0.1159 & 0.0410 & 0.8345 & 0.4353 & 0.3770 & 0.0233 \\
K5, $\alpha$=3.0  & 0.7404 & $-$0.0361 & 0.8505 & 0.1043 & 0.0414 & 0.8529 & 0.4222 & 0.4465 & 0.0196 \\
K10, $\alpha$=1.0 & 0.6080 & $-$0.1685 & 0.7648 & 0.1609 & 0.0305 & 0.7298 & 0.4200 & 0.1369 & 0.0521 \\
K10, $\alpha$=2.0 & 0.6627 & $-$0.1138 & 0.8035 & 0.1368 & 0.0369 & 0.7882 & 0.4229 & 0.2521 & 0.0330 \\
K10, $\alpha$=3.0 & 0.7007 & $-$0.0758 & 0.8296 & 0.1193 & 0.0413 & 0.8219 & 0.4171 & 0.3411 & 0.0249 \\
K20, $\alpha$=1.0 & 0.5381 & $-$0.2383 & 0.7378 & 0.1800 & 0.0286 & 0.6589 & 0.4003 & 0.0591 & 0.0966 \\
K20, $\alpha$=2.0 & 0.6008 & $-$0.1757 & 0.7732 & 0.1576 & 0.0301 & 0.7306 & 0.4071 & 0.1450 & 0.0556 \\
K20, $\alpha$=3.0 & 0.6530 & $-$0.1235 & 0.8077 & 0.1359 & 0.0378 & 0.7828 & 0.4126 & 0.2334 & 0.0360 \\
\bottomrule
\end{tabular}
\end{table}

\section*{95\% confidence intervals on aggregated metrics}

The point estimates reported in the main tables are averages over multiple independent seeds. The 95\% CIs are computed with a two-sided $t$-based method ($t_{n-1}$): $n=6$ seeds for the HEEDB-scale baseline, $n=3$ seeds for AS-norm calibration metrics (Rank@1 is invariant to the seed because the seed only determines the confidence-calibration split, not the reranking), and $n=5$ seeds for GC confidence metrics. In cases where the lower CI bound becomes negative for intrinsically non-negative metrics (e.g., error rates very close to zero), the negative value is an artifact of the $t$-based approximation and should be interpreted as zero.

\begin{table}[ht]
\centering
\caption{HEEDB-scale baseline -- half-width of the 95\% CI over the 84 cells of the grid ($n=6$ seeds). Mean, minimum, and maximum values across all (gallery\_size, exams\_per\_patient) combinations.}
\label{tab:ci_heedb_scale_summary}
\small
\begin{tabular}{lrrr}
\toprule
Metric & Mean half-width & Min & Max \\
\midrule
Rank@1   & 0.0043 & 0.0007 & 0.0143 \\
Rank@5   & 0.0034 & 0.0006 & 0.0138 \\
Rank@10  & 0.0033 & 0.0003 & 0.0132 \\
Rank@100 & 0.0025 & 0.0005 & 0.0109 \\
Rank@200 & 0.0023 & 0.0003 & 0.0087 \\
\bottomrule
\end{tabular}
\par\medskip\footnotesize\textit{Note: the mean half-width of 0.0043 for Rank@1 indicates that the grid estimates are stable across seeds. The cells with the largest half-widths (up to 0.0143) typically correspond to configurations with a low number of exams per patient (Ex=2) and intermediate gallery size, where between-seed sampling variance is naturally higher.}
\end{table}

\begin{table}[ht]
\centering
\caption{AS-norm focus on $N$ -- metrics with 95\% CI ($n=3$ seeds). Rank@1 is invariant to the seed (the seed only determines the calibration split, not the reranking) and is therefore reported as a point estimate. For conf\_acc, Brier, and ECE, values are reported as \textit{estimate} \textsc{[lo, hi]}.}
\label{tab:ci_asnorm_n}
\footnotesize
\begin{tabular}{rp{3.0cm}p{3.8cm}p{3.8cm}p{4.0cm}}
\toprule
$N$ & Rank@1 & conf\_acc@0.5 & conf\_brier & conf\_ece \\
\midrule
1   & 0.7943 & 0.9623 {\scriptsize[0.9574,\,0.9672]} & 0.0391 {\scriptsize[0.0388,\,0.0395]} & 0.0690 {\scriptsize[0.0637,\,0.0742]} \\
2   & 0.7977 & 0.7974 {\scriptsize[0.7934,\,0.8014]} & 0.1580 {\scriptsize[0.1455,\,0.1704]} & 0.0332 {\scriptsize[$-$0.0729,\,0.1394]} \\
3   & 0.8003 & 0.9499 {\scriptsize[0.9441,\,0.9556]} & 0.0653 {\scriptsize[0.0641,\,0.0665]} & 0.1403 {\scriptsize[0.1375,\,0.1432]} \\
4   & 0.8005 & 0.9682 {\scriptsize[0.9657,\,0.9706]} & 0.0536 {\scriptsize[0.0507,\,0.0566]} & 0.1051 {\scriptsize[0.1013,\,0.1089]} \\
5   & 0.7999 & 0.9653 {\scriptsize[0.9635,\,0.9670]} & 0.0482 {\scriptsize[0.0475,\,0.0490]} & 0.0970 {\scriptsize[0.0939,\,0.1000]} \\
6   & 0.7995 & 0.9648 {\scriptsize[0.9613,\,0.9682]} & 0.0460 {\scriptsize[0.0450,\,0.0471]} & 0.0885 {\scriptsize[0.0837,\,0.0932]} \\
7   & 0.7996 & 0.9078 {\scriptsize[0.6695,\,1.1461]} & 0.0840 {\scriptsize[$-$0.0836,\,0.2516]} & 0.0592 {\scriptsize[$-$0.0416,\,0.1600]} \\
8   & 0.7987 & 0.9071 {\scriptsize[0.6684,\,1.1459]} & 0.0837 {\scriptsize[$-$0.0858,\,0.2532]} & 0.0570 {\scriptsize[$-$0.0385,\,0.1524]} \\
9   & 0.7988 & 0.9633 {\scriptsize[0.9578,\,0.9688]} & 0.0442 {\scriptsize[0.0432,\,0.0453]} & 0.0772 {\scriptsize[0.0715,\,0.0828]} \\
10  & 0.7987 & 0.9626 {\scriptsize[0.9578,\,0.9673]} & 0.0443 {\scriptsize[0.0432,\,0.0453]} & 0.0744 {\scriptsize[0.0693,\,0.0795]} \\
\bottomrule
\end{tabular}
\par\medskip\footnotesize\textit{Note: the intervals are computed with a $t$-based approximation over three seeds. When lower bounds become negative or upper bounds exceed 1, this should be interpreted as an artifact of the Gaussian approximation applied to metrics bounded in $[0,1]$. The instability is particularly evident for $N=2$ on ECE and for $N \in \{7,8\}$ on the calibration metrics, indicating high sensitivity of the calibrator to the calibration split in those configurations.}
\end{table}

\begin{landscape}
\begin{table}[p]
\centering
\caption{GC protocol -- retrieval, verification, and confidence metrics.}
\label{tab:gc_conf_full}
\scriptsize
\setlength{\tabcolsep}{3pt}
\begin{tabular}{lrrrrrrrrrrrrrrrrrrrr}
\toprule
DB & Rank@1 & Rank@5 & Rank@10 & TAR@FAR1e-4 & TAR@FAR1e-3 & Acc@0.5 & Brier & ECE & Conf mean & Conf mean y1 & Conf mean y0 & Cov@0.70 & Err@0.70 & Cov@0.80 & Err@0.80 & Cov@0.90 & Err@0.90 & Cov@0.95 & Err@0.95 \\
\midrule
ASUGI & 0.9506 & 0.9670 & 0.9718 & 0.9448 & 0.9675 & 0.9713 & 0.0207 & 0.0185 & 0.9461 & 0.9702 & 0.4820 & 0.9335 & 0.0076 & 0.9149 & 0.0040 & 0.8805 & 0.0014 & 0.8369 & 0.0009 \\
MIMIC & 0.8291 & 0.8768 & 0.8916 & 0.7079 & 0.8188 & 0.8756 & 0.0849 & 0.0340 & 0.8278 & 0.8916 & 0.5185 & 0.7940 & 0.0567 & 0.7236 & 0.0349 & 0.5889 & 0.0166 & 0.4175 & 0.0090 \\
HEEDB & 0.6884 & 0.7593 & 0.7807 & 0.6002 & 0.7210 & 0.8366 & 0.1185 & 0.0286 & 0.6865 & 0.8173 & 0.3969 & 0.5835 & 0.0727 & 0.4992 & 0.0528 & 0.3573 & 0.0331 & 0.2269 & 0.0252 \\
\bottomrule
\end{tabular}
\end{table}

\begin{table}[p]
\centering
\caption{GC confidence scores -- main metrics with 95\% CI ($n=5$ seeds). Format: \textit{estimate} {\scriptsize[lo, hi]}.}
\label{tab:ci_gc_main}
\footnotesize
\begin{tabular}{lp{3.2cm}p{3.2cm}p{3.2cm}p{3.2cm}p{3.2cm}p{3.2cm}}
\toprule
Dataset & Acc@0.5 & Brier & ECE & $\bar{c}$ & $\bar{c}_{y=1}$ & $\bar{c}_{y=0}$ \\
\midrule
ASUGI-DB
  & 0.9991 {\scriptsize[0.9991,\,0.9992]}
  & 0.0008 {\scriptsize[0.0007,\,0.0009]}
  & 0.0058 {\scriptsize[0.0050,\,0.0066]}
  & 0.9939 {\scriptsize[0.9935,\,0.9943]}
  & 0.9940 {\scriptsize[0.9936,\,0.9943]}
  & 0.9226 {\scriptsize[0.8856,\,0.9595]} \\[4pt]
MIMIC-GC
  & 0.9733 {\scriptsize[0.9727,\,0.9739]}
  & 0.0173 {\scriptsize[0.0171,\,0.0175]}
  & 0.0216 {\scriptsize[0.0214,\,0.0218]}
  & 0.9574 {\scriptsize[0.9569,\,0.9580]}
  & 0.9729 {\scriptsize[0.9725,\,0.9732]}
  & 0.5481 {\scriptsize[0.5413,\,0.5550]} \\[4pt]
HEEDB-GC
  & 0.9652 {\scriptsize[0.9645,\,0.9659]}
  & 0.0250 {\scriptsize[0.0248,\,0.0252]}
  & 0.0250 {\scriptsize[0.0243,\,0.0256]}
  & 0.9370 {\scriptsize[0.9362,\,0.9377]}
  & 0.9631 {\scriptsize[0.9626,\,0.9637]}
  & 0.5022 {\scriptsize[0.4953,\,0.5091]} \\
\bottomrule
\end{tabular}
\par\medskip\footnotesize\textit{Note: the wide CI of $\bar{c}_{y=0}$ on ASUGI-DB [0.886, 0.960] reflects the variability of scores on the rare incorrect cases (Rank@1 = 0.9506); the absolute number of errors is low and the conditional estimate is intrinsically less stable.}
\end{table}

\begin{table}[p]
\centering
\caption{GC confidence scores -- coverage/error metrics with 95\% CI ($n=5$ seeds). Format: \textit{estimate} {\scriptsize[lo, hi]}.}
\label{tab:ci_gc_cov_err}
\footnotesize
\begin{tabular}{lp{2.8cm}p{2.8cm}p{2.8cm}p{2.8cm}p{2.8cm}p{2.8cm}p{2.8cm}p{2.8cm}}
\toprule
Dataset & Cov@0.70 & Cov@0.80 & Cov@0.90 & Cov@0.95 & Err@0.70 & Err@0.80 & Err@0.90 & Err@0.95 \\
\midrule
ASUGI-DB
  & 1.0000 {\scriptsize[1.000,\,1.000]}
  & 1.0000 {\scriptsize[1.000,\,1.000]}
  & 0.9994 {\scriptsize[0.9986,\,1.000]}
  & 0.9973 {\scriptsize[0.9951,\,0.9995]}
  & 0.0009 {\scriptsize[0.0008,\,0.0009]}
  & 0.0009 {\scriptsize[0.0008,\,0.0009]}
  & 0.0006 {\scriptsize[0.0002,\,0.0010]}
  & 0.0003 {\scriptsize[$-$0.0000,\,0.0007]} \\[4pt]
MIMIC-GC
  & 0.9504 {\scriptsize[0.9490,\,0.9519]}
  & 0.9324 {\scriptsize[0.9317,\,0.9332]}
  & 0.9000 {\scriptsize[0.8991,\,0.9008]}
  & 0.8528 {\scriptsize[0.8514,\,0.8541]}
  & 0.0058 {\scriptsize[0.0047,\,0.0069]}
  & 0.0018 {\scriptsize[0.0016,\,0.0020]}
  & 0.0005 {\scriptsize[0.0005,\,0.0005]}
  & 0.0003 {\scriptsize[0.0003,\,0.0003]} \\[4pt]
HEEDB-GC
  & 0.9171 {\scriptsize[0.9164,\,0.9179]}
  & 0.8945 {\scriptsize[0.8937,\,0.8954]}
  & 0.8529 {\scriptsize[0.8520,\,0.8538]}
  & 0.8016 {\scriptsize[0.8003,\,0.8028]}
  & 0.0065 {\scriptsize[0.0062,\,0.0068]}
  & 0.0034 {\scriptsize[0.0033,\,0.0035]}
  & 0.0018 {\scriptsize[0.0017,\,0.0019]}
  & 0.0012 {\scriptsize[0.0011,\,0.0013]} \\
\bottomrule
\end{tabular}
\par\medskip\footnotesize\textit{Note: the very tight CIs for MIMIC and HEEDB on Err@0.90 and Err@0.95 indicate that high-threshold error estimates are very stable across seeds; the only CI with a negative lower bound (Err@0.95 ASUGI) is a $t$-based artifact and should be read as zero.}
\end{table}
\end{landscape}